%% file: main.tex
\documentclass[sigconf, nonacm]{acmart}

\newcommand\vldbdoi{XX.XX/XXX.XX}
\newcommand\vldbpages{XXX-XXX}
\newcommand\vldbvolume{14}
\newcommand\vldbissue{1}
\newcommand\vldbyear{2020}
\newcommand\vldbauthors{\authors}
\newcommand\vldbtitle{\shorttitle} 
\newcommand\vldbavailabilityurl{https://github.com/UIC-InDeXLab/Chameleon}
\newcommand\vldbpagestyle{plain} 

\input{header}

\begin{document}
% \title{Foundation Models for Fairness-aware Data Augmentation: \system for Coverage Enhancement in Image Datasets}
% \title{\system: Foundation Models for Fairness-aware Data Augmentation to Enhance Coverage in Image Datasets}
\title{\system: Foundation Models for Fairness-aware Multi-modal Data Augmentation to Enhance Coverage of Minorities}

\author{Mahdi Erfanian}
\orcid{0009-0008-2526-0470}
\affiliation{%
  \institution{University of Illinois Chicago}
}
\email{merfan2@uic.edu}

\author{H. V. Jagadish}
\orcid{0000-0003-0724-5214}
\affiliation{%
  \institution{University of Michigan}
}
\email{jag@umich.edu}

\author{Abolfazl Asudeh}
\orcid{0000-0002-5251-6186}
\affiliation{%
  \institution{University of Illinois Chicago}
}
\email{asudeh@uic.edu}

\begin{abstract}
The potential harms of the under-representation of minorities in training data, particularly in multi-modal settings, is a well-recognized concern.
While there has been extensive effort in detecting such under-representation,  resolution has remained a challenge.

With recent advancements in generative AI, large language models and foundation models have emerged as versatile tools  across various domains.
In this paper, we propose \system, a system that efficiently utilizes these tools to augment a data set with a minimal addition of synthetically generated tuples, in order to enhance the coverage of the under-represented groups. Our system follows a rejection sampling approach to ensure the generated tuples have a high quality and follow the underlying distribution.
In order to minimize the rejection chance of the generated tuples, we propose multiple strategies for providing a guide for the foundation model.
Our experiment results, in addition to confirming the efficiency of our proposed algorithms, illustrate the effectiveness of our approach, as the unfairness of the model in a downstream task significantly dropped after data repair using \system. 
\end{abstract}

\maketitle

%%% do not modify the following VLDB block %%
%%% VLDB block start %%%
\pagestyle{\vldbpagestyle}
\begingroup\small\noindent\raggedright\textbf{PVLDB Reference Format:}\\
\vldbauthors. \vldbtitle. PVLDB, \vldbvolume(\vldbissue): \vldbpages, \vldbyear.\\
\href{https://doi.org/\vldbdoi}{doi:\vldbdoi}
\endgroup
\begingroup
\renewcommand\thefootnote{}\footnote{\noindent
This work is licensed under the Creative Commons BY-NC-ND 4.0 International License. Visit \url{https://creativecommons.org/licenses/by-nc-nd/4.0/} to view a copy of this license. For any use beyond those covered by this license, obtain permission by emailing \href{mailto:info@vldb.org}{info@vldb.org}. Copyright is held by the owner/author(s). Publication rights licensed to the VLDB Endowment. \\
\raggedright Proceedings of the VLDB Endowment, Vol. \vldbvolume, No. \vldbissue\ %
ISSN 2150-8097. \\
\href{https://doi.org/\vldbdoi}{doi:\vldbdoi} \\
}\addtocounter{footnote}{-1}\endgroup
%%% VLDB block end %%%

%%% do not modify the following VLDB block %%
%%% VLDB block start %%%
\ifdefempty{\vldbavailabilityurl}{}{
\vspace{.3cm}
\begingroup\small\noindent\raggedright\textbf{PVLDB Artifact Availability:}\\
The source code, data, and/or other artifacts have been made available at \url{\vldbavailabilityurl}.
\endgroup
}
%%% VLDB block end %%%

\input{intro}

\input{pre}
\input{rejectionsampling}
\input{analyzer}

\input{selection}
\input{exp}

\input{related}
\input{proofs}
\input{conclusion}
\newpage
% \begin{acks}
%  This work was supported by the [...] Research Fund of [...] (Number [...]). Additional funding was provided by [...] and [...]. We also thank [...] for contributing [...].
% \end{acks}

%\clearpage

\bibliographystyle{ACM-Reference-Format}
\bibliography{ref}
\end{document}

%% file: header.tex
\usepackage{flushend}
\usepackage[a-1b]{pdfx}   % for PDF/A-1b
\usepackage{balance}

\usepackage{color}
\usepackage{tcolorbox}
\usepackage[labelfont=bf]{caption}
\usepackage{subcaption}
\usepackage{hyperref}
\usepackage{makecell}
\usepackage{fixltx2e}
\usepackage{rotating,multirow}
\usepackage{etoolbox,xspace}
\usepackage{algorithmicx}
\usepackage{algorithm}
\usepackage{fontawesome}
\usepackage{algpseudocode}
\usepackage{wrapfig}
\usepackage{graphicx}

\algtext*{EndWhile}
\algtext*{EndIf}
\algtext*{EndFunction}
\algtext*{EndFor}
\algrenewcommand\algorithmicrequire{\textbf{Input:}}
\algrenewcommand\algorithmicensure{\textbf{Output:}}

\usepackage{amsthm}
\DeclareMathOperator*{\argmax}{argmax} 
\usepackage{tabularx}
\usepackage{ragged2e}  % for '\RaggedRight' macro (allows hyphenation)
\usepackage{booktabs}  % for \toprule, \midrule, and \bottomrule macros
\usepackage{textcomp}
\usepackage{url}
\usepackage{comment}
\usepackage{enumitem}
\usepackage[simplified]{pgf-umlcd}
\usepackage{tikz}
\usepackage{pgfplots}
\usepackage{subcaption}

\usepackage{filecontents}

\input{plots/l2-200}
\input{plots/l2-350}
\input{plots/l1-1000}
\input{plots/l1-2000}
\input{plots/feret-lack-of-coverage}
\newcolumntype{Y}{>{\RaggedRight\arraybackslash}X}

\newtheorem{theorem}{Theorem}
\newtheorem{lemma}{Lemma}

\newtcolorbox{defbox}{
  colback=orange!10,
  colframe=orange!20,
  arc=2mm, % Rounder edges
  fonttitle=\bfseries,
  boxrule=0mm,
  boxsep=1mm,
  left=0mm,
  right=0mm,
  top=0mm,
  bottom=0mm
}

\newcommand*{\numberingI}[1]{%
\footnotesize\protect\tikz[baseline=-3px]%
\protect\node[draw,circle,inner sep=1pt](n1){#1};}

\newcounter{example}%[section]
\renewcommand{\theexample}{\arabic{example}}

\newtcolorbox{examplebox}{
  colback=blue!10,
  colframe=blue!20,
  arc=2mm, % Rounder edges
  fonttitle=\bfseries,
  boxrule=0mm,
  boxsep=1mm,
  left=0mm,
  right=0mm,
  top=0mm,
  bottom=0mm
}

\newenvironment{proof}{\paragraph{Proof:}}{\hfill$\square$}
\newcommand{\squishlist}{
 \begin{list}{$\bullet$}
  { \setlength{\itemsep}{0pt}
     \setlength{\parsep}{1pt}
     \setlength{\topsep}{1pt}
     \setlength{\partopsep}{0pt}
     \setlength{\leftmargin}{1em}
     \setlength{\labelwidth}{1em}
     \setlength{\labelsep}{0.5em} } }

\newcommand{\squishend}{
  \end{list}
}
\definecolor{americanrose}{rgb}{1.0, 0.70, 0.70}
\definecolor{airforceblue}{rgb}{0.70, 0.70, 1.00}
\definecolor{redouter}{rgb}{1.0, 0.19, 0.19}
\definecolor{blueouter}{rgb}{0.23, 0.23, 1.0}
\definecolor{ao(english)}{rgb}{0.0, 0.5, 0.0}
\definecolor{ao}{rgb}{0.0, 0.0, 1.0}

\newcommand{\mahdi}[1]{\textcolor{red}{Mahdi: #1}}

\newcommand{\eat}[1]{}

\usepackage{xspace}

\newcommand{\at}[1]{{\tt \small #1}\xspace}
\newcommand{\system}{{\sc Chameleon}\xspace}
\newcommand{\fm}{$\mathcal{F}$\xspace}
\newcommand{\dalle}{{\sc DALL·E 2}\xspace}
\newcommand{\similarity}{\mathcal{S}_{im}}

\newcommand{\greedy}{{\sc Greedy}\xspace}
\newcommand{\linucb}{{\sc LinUCB}\xspace}
\newcommand{\cs}{{\sc Combination-Selection}\xspace}

\newcommand{\vc}{{\sc VC}\xspace}
\newcommand{\opt}{{\sc OPT}\xspace}

\newcommand{\utkface}{{\sc UTKFace}\xspace}
\newcommand{\feretdb}{{\sc FERETDB}\xspace}

\newcommand{\gee}{\mathbf{g}}
\newcommand{\dee}{\mathcal{D}}
\newcommand{\dist}{\xi}

\newcommand{\np}{{\sc NP}}

%% file: intro.tex
\section{Introduction}\label{sec:intro}

{\em ``The \system changes color to match the earth, the earth doesn't change color to match the chameleon.''}
\vspace{-4mm}
\begin{flushright}
-- \textsc{Senegalese Proverb}
\end{flushright}
\vspace{1mm}
\noindent

With the wide use of machine learning, the importance of using appropriate data sets is now well-recognized. In particular, there is increasing awareness of unfairness towards minorities and other marginalized groups on account of their under-representation in the training data.  There is now a growing body of work on {\em coverage} in a training data set \cite{shahbazi2023representation, mousavi2024data,asudeh2019assessing}, giving us tools to detect such under-representation.  

Of course, detecting under-representation does not in itself address the problem: we then need to fix it somehow, for example by integrating data from external sources~\cite{nargesian2021tailoring}.  If additional data could be collected, that would be ideal, but this is frequently not possible.  An approach in such cases is to generate synthetic data.  Indeed, such approaches have been explored for regular alphanumeric data, such as in relational tables~\cite{DBLP:journals/jair/ChawlaBHK02,DBLP:conf/icic/HanWM05,fan2020relational,cai2023privlava}.

Multi-modal data is increasingly being used for analysis, exploiting huge recent advances in technologies such as image recognition.  In fact, bias in multi-modal training data has been noticed for quite some time, beginning with the well-know case of early google software labeling an African American woman as a gorilla~\cite{google-gorilla}, and continuing through many other cases~\cite{buolamwini2018gender,langston2015ceo,closed-eyes,hp1}.  This begs the question, what can we do once we have detected that a multi-modal dataset is biased, with insufficient representation of certain groups?  There is no obvious way we can apply techniques developed for alphanumeric relational data.  

This is the problem we tackle in this paper.  Our central idea is to use generative AI to create synthetic data for this purpose.  While this idea has immediate appeal, particularly given the spectacular recent advances in Foundation Models, actually getting it to work requires overcoming many challenges.  
First, we have to determine the minimal set of synthetic tuples that once added to the original data set, under-representation issues are resolved. % what we should ask the foundation model to generate in order resolve under-representation issues with minimum synthetic data augmented into the input data set.
% AI-image generator, we have to check whether the generated images are actually reasonable, and so on.  We also have to worry about distribution shifts due to synthetically generated images, and other subtle problems we discuss below.
% \jag{This paragraph needs to be rewritten and extended once the challenge statement (in the new section 3) is clear.}
% There are multiple challenges towards effective data set augmentations using foundation models. 
Second, in order to prevent a distribution shifts due to synthetically generated tuples, we need to ensure that the generated data follows the underlying distribution represented by the input data set.
Third, we have to ensure the generated images are actually reasonable and have a high quality to look realistic to a human evaluator. 
Last but not least, given the (often monetary) cost associated with the queries to the foundation model, 
we should ensure the cost-effectiveness of the data set repair process.

To address the first challenge, using the notion of data coverage~\cite{asudeh2019assessing,shahbazi2023representation} for identifying under-representation, we formally define the \cs problem, which minimizes the total number of synthetic tuples for resolving lack of coverage of minorities at the most general level. We show the problem is {\sc NP}-hard, and propose a greedy approximation algorithm for it.
For the second challenge, we view each tuple in the data set $\dee$ as an independent and identically distributed (iid) random sample from the underlying distribution $\xi$ it represents. We use the vector representations (embeddings) space to describe the distribution. Then, a newly generated tuple is discarded if it fails the data distribution test, i.e., if it is unlikely to be generated by $\xi$.
To address the third challenge, we model the quality evaluation as hypothesis testing, and reject the samples that have a higher chance of being labeled as ``unrealistic'' by a random human evaluator.
Finally, to minimize the number of queries to the foundation model, we provide a guide tuple (and a mask), in addition to the prompt, to the foundation model. 
We propose multiple strategies for guide selection such that the chance of passing the distribution and the quality tests is maximized.

\paragraph{Summary of contributions}
We introduce \system, a system that uses foundation models to augment multi-modal data sets in order to enhance their representation of the minorities, in form of data coverage. 
To the best of our knowledge, our paper is {\em the first to use foundation models for fairness-aware data augmentation}.
In summary, our contributions are the following:

\begin{itemize}[leftmargin=*]
    \item We propose fairness-aware data augmentation using foundation models for resolving lack of coverage in multi-modal data (\S~\ref{sec:pre}).
    \item Following a rejection sampling approach, we propose data distribution and quality evaluation tests to ensure the augmented tuples do not deviate from the underlying data distribution, and have as high quality as the real tuples in the data set (\S~\ref{sec:rejectionsampling}).  
    \item We propose the \cs problem, which specifies the description of the tuples to be generated with the goal to resolve lack of coverage with minimum amount of augmentation to the data set. We prove that the \cs problem is \np-hard, and propose a greedy approximation algorithm with the logarithmic approximation-ratio for it (\S~\ref{sec:analyzer}).
    \item We propose the Guide-selection problem that provides a guide tuple and a mask as the input to the foundation model in order to maximize the chance of passing the rejection sampling tests. We propose multiple strategies for guide-selection, including a solution based on contextual multi-armed bandit (\S~\ref{sec:selection}).
    \item We conduct comprehensive experiments on real and synthetic data sets to evaluate the efficiency of the proposed  algorithms in comparison to the baselines and to study their effectiveness using human evaluators.
    We also provide an experiment that illustrates the unfairness (performance disparity) of a CNN model on uncovered groups significantly decreases when trained on a repaired data set using \system (\S~\ref{sec:exp}).
\end{itemize}

%% cut for space limitations
% \paragraph{Paper organization}
% The rest of the paper is organized as following.
% First in \S~\ref{sec:pre}, we define the terms and notations, clarify the inputs and the objective, and provide an overview of our system.
% Then in \S~\ref{sec:rejectionsampling}, we propose our rejection sampling approach for the data distribution and quality evaluation tests. 
% The \cs problem is studied in \S~\ref{sec:analyzer}. We then propose the guide-selection problem in \S~\ref{sec:selection} and propose multiple approaches for it. The experiments are provided in \S~\ref{sec:exp}, followed by related work, proofs, and final remarks in \S~\ref{sec:related}, \S~\ref{sec:proofs}, and \S~\ref{sec:conclusion}.

% challenges:
% respecting the underlying distribution
% output quality.
% budget constraint.

%ideas to overcome the challenges
% Identifying the underlying distribution is cumbersome

% instead we make the observation that the dataset at hand is expected to be an unbissed collection of samples, following the underlying distribution. That is, every sample is an iid sample from the unknown underlying distribution.
% in such a setting, a random sample from the data set is an unbiased sample from the underlying distribution.

% using this observation, our idea is to select a random sample from the underlying distribution and use it to "guide" the image generation process.

% for ensuring the output quality, human evaluators can be used. 

% now switch to the use-case: image data sets

%% file: pre.tex
\section{Preliminaries}\label{sec:pre}
\subsection{(Input) Data Model}
% data in form of a set of objects
We are given a data set of multi-modal tuples (e.g., images) $\dee=\{t_1,\cdots, t_n\}$, as a collection
of independent and identically distributed (iid) samples, taken from an (unknown) distribution $\dist$. 
The tuples are associated with $d\geq 1$ attributes of interest $\mathbf{x}=\{x_1,\cdots,x_d\}$
(e.g., \at{gender}, \at{race}, \at{age-group}, etc.), that are used for identifying (demographic) groups.
Without loss of generality, we assume the attributes of interest are categorical (we assume the continuous attributes are properly bucketized).
Attributes of interest can be unordered (e.g., \at{gender} and \at{race}) or ordinal (e.g., \at{age-group}).
% While some attributes of interest such as \at{gender} and \at{race} are non-ordinal, others such as \at{age-group} are ordinal.
% We can divide all attributes to two sets of \textit{ordered} and \textit{non-ordered} attributes. Ordered attributes are attributes that by nature, have a proceeding logical order within. eg: \texttt{age\_group} is an ordered attribute because it has an order within, it starts from the infant and goes to the elderly. 
% Non-ordered attributes do not have any logical order, for example, \textit{race} is a non-ordered attribute because there is no logical order in the list of acceptable values for race. \texttt{[White, Black, Asian, Indian]}.
Each attribute has a cardinality of two or more. %, specifying different non-overlapping (demographic) groups.
For example, an attribute {\small \tt sex} (biological sex) with values {\small \tt \{male, female\}} partitions the individuals into two non-overlapping groups.
We use $dom(x_i)$ to represent the 
domain of the attribute $x_i\in \mathbf{x}$, i.e., the set of valid values for $x_i$.
% cardinality of the attribute $x_i\in \mathbf{x}$.

The cartesian product of values on a subset of attributes $\mathbf{x}'\subseteq \mathbf{x}$, form a set of (demographic) subgroups.
For example, $\{$\at{white male}, \at{white female}, \at{black male} $,\cdots\}$ are the subgroups defined on the attributes \at{(race,gender)}.
We refer to the number of attributes used to specify a subgroup as the {\em level} of that subgroup.
For example, the level of the subgroup \at{white male} is 2, while the level of the subgroup \at{male} is 1.
We use $\ell(\gee)$, to refer to the level of a subgroup $\gee$.
Similarly, we say a subgroup $\gee'$ is a subset of $\gee$, if the groups specifying $\gee'$ are a superset of the ones for $\gee$. For example \at{\{white male preschooler\}} is a subset of the more general group \at{\{white male\}}. That is, the set of individuals in group \at{\{white male preschooler\}} are a subset of \at{\{white male\}}.
Moreover, we say a subgroup $\gee$ is a {\em parent} of the subgroup $\gee'$, if $\gee'\subset \gee$ and $\ell(\gee)=\ell(\gee')+1$. For example, the subgroup \at{\{white male\}} is a parent of the subgroup \at{\{white male preschooler\}}.
Finally, slightly abusing the terms, we call a subgroup a {\em combination} if  $\ell(\gee) = d$.
Furthermore, we say two combinations are {\em ``sibling''} if they differ in exactly one attribute (with their values the same on all other attributes).

% Furthermore, we introduce the concept of a combination $c$, defined as a pattern devoid of any unknown attributes. Formally, a subgroup pattern qualifies as a combination if and only if $\ell(\gee) = d$. 
\input{ToN}
\subsection{(Input) Foundation Model}\label{sec:pre:foundationmodel}
We assume access to a foundation model \fm (e.g., \dalle\footnote{ \system uses \dalle as its default image generator. ``\dalle is an AI system that can create realistic images and art from a description in natural language.'' \url{https://openai.com/dall-e-2}}) for data generation.
In the following, we discuss the requirements relevant to this project. 
We treat \fm as black-box, which allows the adaptation of both closed-source and open-source foundation models.
For more information about the foundation models, please refer to \cite{zhou2023comprehensive, bubeck2023sparks,bommasani2021opportunities}.
We consider the foundation model \fm with the following inputs that generates a synthesized output tuple:
\begin{itemize}[leftmargin=*]
    \item {\bf Prompt}: a natural language description of an instruction that specifies the details of the tuple to be generated. For example, a prompt for image generation could be ``A realistic photo of a white cat''.
    \item {\bf Guide}: when only provided with a prompt, the foundation model uses its "imagination" to generate the requested tuple. For example, for the previous cat-image example prompt, the breed and size of the cat, the background, and other details are chosen by the foundation model. Alternatively, a guide can be provided to \fm to
    influence the generation process. We formalize the guide as a pair $(t,m)$, where $t$ is a tuple, and $m$ is a mask. The mask $m$ specifies which parts of the guide tuple should change. Continuing with the cat example, $t$ can be a cat image and $m$ can specify the foreground to be regenerated.
\end{itemize}

\paragraph{Cost model:} We assume each query to the foundation model has a fixed cost $\upsilon$. The cost is monetary when using external foundation models such as \dalle, and it can be computational when the model is hosted locally.

% The LLM is randomized, i.e., the tokens are sequentially drawn based on the underlying distribution of the (top-k or top-p\%) token-probabilities. We treat the LLM as a black-box oracle that upon querying generates an {\em output} based on the input prompt.

\subsection{(Objective) Data Coverage}\label{sec:pre:datacoverage}

We use the notion of {\em data coverage} \cite{asudeh2019assessing} to identify lack of representation issues in a dataset $\dee$.
In particular, given a data set $\dee$ and a coverage threshold $\tau$ (e.g., $\tau=50$), we say a subgroup $\gee$ is {\em uncovered}, if $|\gee\cap\dee|<\tau$. That is, the number of samples in $\dee$ from the group $\gee$ are less than $\tau$.
When studying lack of coverage in a data set, we are usually interested in finding the most-general uncovered subgroups. That is, the collection of subgroups $\gee$ such that (a) $\gee$ is uncovered and (b) all parents of $\gee$ are covered.

Following the notation in \cite{asudeh2019assessing}, we use \textit{patterns} to refer to uncovered subgroups.
A pattern $P$ is a string of $d$ values, where $P[i]$ is either a value from the domain of $x_i$, or it is ``unspecified'', specified with $X$. 
For example, consider a dataset with three binary attributes of interest $\mathbf{x}=\{x_1, x_2, x_3\}$. The pattern $P=X01$ specifies all the tuples for which $x_2=0$ and $x_3=1$ ($x_1$ can have any value).
The set of patterns that identify most-general uncovered subgroups are called {\em Maximal Uncovered Patterns} (MUPs).
The level of a MUP is the same as the level of the subgroup it represents. 
% \jag{While the rest of this paragraph is accurate, let us delete unless necessary for this paper?} While MUPs at the higher levels (small subgroups that are a combination of several groups) are generally less problematic, the MUPs with small levels are troublesome.
% Our objective with augmenting a multi-modal data set is to (at least) resolve the lack of coverage for the small levels.

\subsection{System Architecture Overview}
\begin{figure}[!t]
\begin{tcolorbox}[
    colback=gray!10,
    colframe=gray!10,
    boxrule=0.5pt,
    rounded corners,
    boxsep=0pt,
    width=0.5\textwidth
]
    \centering
    \includegraphics[width=\textwidth]{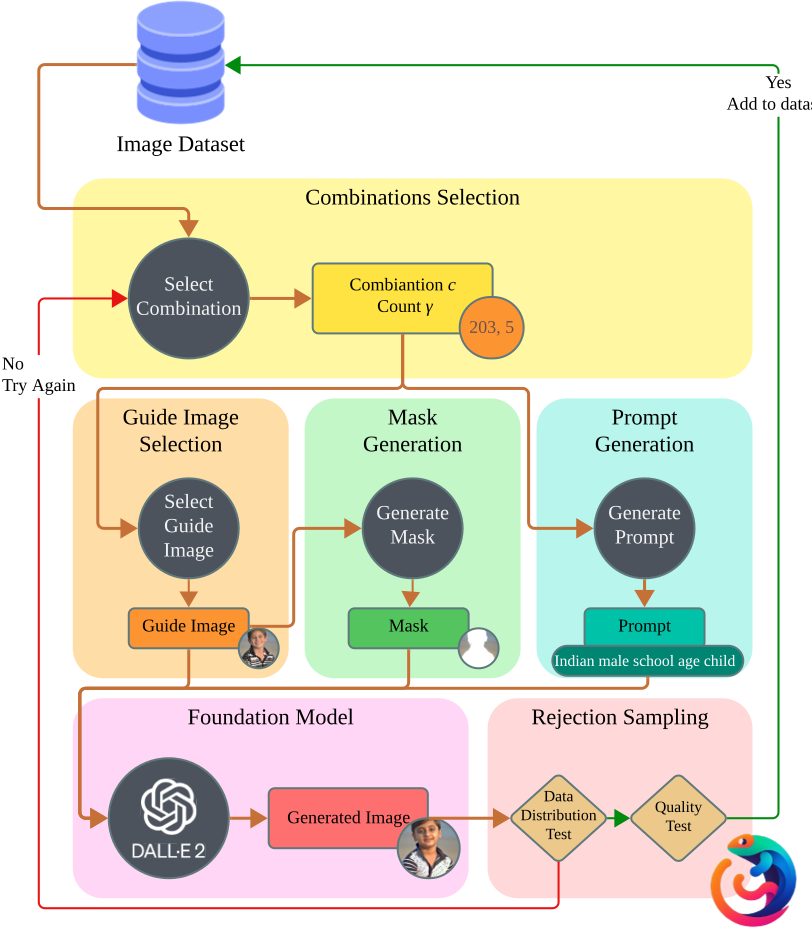}
    \vspace{-4mm}
    \caption{Architecture of \system for image generation}
    \label{fig:pipeline}
\end{tcolorbox}
\vspace{-5mm}
\end{figure}

% In this section, we briefly go over different part of 
%Before starting the technical discussions, let us provide a high-level overview of the architecture of \system.
Figure~\ref{fig:pipeline} shows the overall architecture of the system, whose components we will design in the rest of this paper. 
The augmentation process for a data set $\dee$ starts with specifying a small set of synthetic tuples (a set of combinations, each with a count) that once generated and added to $\dee$, resolve problematic lack of coverage issues (\S~\ref{sec:analyzer}).
% Then for each combination, an input query is constructed to be passed to the foundation model. Specifically, in addition to a prompt, a guide tuple (and mask) is selected to maximize the chance of generating a high quality tuple that follows the underlying distribution of $\dee$ (\S~\ref{sec:selection}).
% The query is then passed to the foundation model to generate a new tuple.
% Then, following a rejection sampling strategy, the new tuple should pass a data distribution test and a quality test before it can be augmented to the dataset (\S~\ref{sec:rejectionsampling}).
% \jag{The above is too cryptic. 
%  I have tried to explain below.  Please check what I wrote, and correct as needed.}
Then for each combination, an input query is constructed to be passed to the foundation model. At minimum, this query comprises a text prompt describing the desired combination. However, that leaves too much latitude to the foundation model, and the result is likely to be an image unsuitable for the data set at hand (that is, it would be unlikely to occur in the underlying distribution of $\dee$), even if it satisfies the prompt conditions. To avoid this, a guide tuple, and mask, is specified, in addition to the prompt (\S~\ref{sec:selection}).
The foundation model then generates a new tuple based on this input. Even with the augmented query, the produced tuple may not be satisfactory.
We follow a rejection sampling strategy: the new tuple should pass a data distribution test and a quality test before it can be added to the dataset (\S~\ref{sec:rejectionsampling}).

%% file: ToN.tex
% \begin{table}[t]
% \caption{Table of Notations}\label{tab:notations}
% \begin{tabular}{|l|l|}
% \hline
% {\bf notation} & {\bf description} \\ \hline
% $\dee=\{t_1\cdots,t_n\}$&the data set\\ \hline 
% $\mathbf{x}=\{x_1\cdots,x_d\}$&attributes of interest\\ \hline
% $dom(x_i)$&the set of valid values for the attribute $x_i$\\ \hline
% $\ell(\gee)$&the level of a subgroup $\gee$\\ \hline
% $\tau$&the coverage threshold\\ \hline
% $P_i$&a pattern, specifying a subgroup over $\mathbf{x}$\\ \hline
% $c_i$&a combination in $ \varprod_{k=1}^d dom(x_k)$\\ \hline
% \fm& the foundation model\\ \hline
% $\upsilon$&the query cost of \fm\\ \hline
% $\dist$& the underlying distribution of $\dee$\\ \hline
% $\vec{v}(t_i)$& the vector representation of $t_i$\\ \hline
% \end{tabular}
% \end{table}
\begin{table}[t]
    \centering
    \caption{Table of Notations}\label{tab:notations}
    \vspace{-3mm}
    \begin{tabular}{ll}
        \toprule
        \textbf{Notation} & \textbf{Description} \\
        \midrule
        $\dee=\{t_1\cdots,t_n\}$ & the data set \\
        $\mathbf{x}=\{x_1\cdots,x_d\}$ & attributes of interest \\
        $dom(x_i)$ & the set of valid values for the attribute $x_i$ \\
        $\ell(\gee)$ & the level of a subgroup $\gee$ \\
        $\tau$ & the coverage threshold \\
        $P_i$ & a pattern, specifying a subgroup over $\mathbf{x}$ \\
        $c_i$ & a combination in $\varprod_{k=1}^d dom(x_k)$ \\
        \fm & the foundation model \\
        $\upsilon$ & the query cost of \textit{FM} \\
        $\dist$ & the underlying distribution of $\dee$ \\
        $\vec{v}(t_i)$ & the vector representation of $t_i$ \\
        \bottomrule
    \end{tabular}
\end{table}

%% file: rejectionsampling.tex
\section{Rejection Sampling}\label{sec:rejectionsampling}
Our strategy for ensuring the high quality of the augmented dataset is inspired by rejection sampling~\cite{hammersley2013monte,flury1990acceptance}. In order to generate a sample from a distribution with the probability density function (pdf) $f$, the rejection sampling technique
generates sample points under an upper-envelope of $f$ and rejects it if the sample point does not fall under $f$.
% considers an upper-envelop over $f$ and draws a sample point uniformly at random from the area under the upper-envelop. Next, it accepts (outputs) the x-value of the generated point, if it falls under $f$, and rejects its otherwise.
We follow a similar strategy. Specifically, when a new tuple is generated by the foundation model, we only accept it if it passes the data distribution test (\S~\ref{sec:test1}) and the quality evaluation (\S~\ref{sec:test2}). Otherwise, the generated tuple is rejected and we try again.

\subsection{Data distribution test}\label{sec:test1}
When augmenting a data set, it is critical to ensure that added tuples follow the underlying data distribution $\dist$. %\nima{the previous sentence reads a little informal to me.}
For example, when the data set comprises wide-shot images at an office workplace the generated tuples should also belong to the same context.
The first issue though is that $\dist$ is unknown. Besides, it is not clear how to quantify and represent the distribution, while
relying on the foundation model's imagination could cause a distribution drift.

We utilize the vector representation (aka embedding) of the tuples for representing the distribution $\dist$.
Given a tuple $t_i$, let $\vec{v}(t_i) = \vec{v}_i = \langle v_1, v_2, \cdots, v_\kappa\rangle$ be its embedding.
We assume the embeddings are accurate. That is, the cosine similarity between the embeddings represent the semantic similarity between two tuples. Formally, the similarity of two tuples $t_i$ and $t_j$ can be computed as $\similarity(t_i,t_j) = \cos{\angle(\vec{v}_i,\vec{v}_j)}$.
% Similarly, the distance between $t_i$ and $t_j$ is defined as $\distance(t_i,t_j) = 1-\similarity(t_i,t_j)$.
Now, in the embedding space 
let $\dist$ be the probability distribution 
from which $\dee$ is sampled.
Hence, the probability that a tuple $t$ is sampled is $Pr_\dist(t)$. 
We use $\vec{\mu}_\dist$ to represent the mean of $\dist$.
Since the tuples in $\dee$ are iid samples from $\dist$, those can be used for estimating $\vec{\mu}_\dist$. Let $\vec{v}_c$ be the sample mean of the representation vectors in $\dee$. That is,
$\vec{v}_c = \frac{1}{m}\sum_{t_i\in\dee} \vec{v}_i$.
Assuming that $n$ (the size of $\dee$) is large enough, based on the central limit theorem we can estimate $\vec{\mu}_\dist$ as $\vec{v}_c$. 

To ensure that generated tuples adhere to the underlying distribution $\dist$, we employ a one-class support vector machine (OCSVM) approach proposed by Scholkopf et al.~\cite{scholkopf1999support} as a quality control mechanism.

Formally, given a set of training embeddings $\{\vec{v}_1, \vec{v}_2, \ldots, \vec{v}_n\}$ representing tuples drawn from $\dist$, the OCSVM aims to learn a decision boundary that separates the majority of these embeddings from the origin in the feature space. This boundary implicitly defines a region that characterizes the "normal" or acceptable embeddings.
To find this boundary (hyperplane), the
following optimization problem is proposed:

\vspace{-5mm}
\begin{align*}
\min_{\textbf{w}, \rho, \textbf{$\epsilon$}} \quad & \frac{1}{2}||\textbf{w}||^2 + \frac{1}{\nu n}\sum_{i=1}^n \epsilon_i - \rho \\
\text{Subject to} \quad & \textbf{w} \cdot \phi(\vec{v}_i) \geq \rho - \epsilon_i, \quad \epsilon_i \geq 0, \quad i = 1, 2, \ldots, n
\end{align*}

\noindent where:
\begin{itemize}[leftmargin=*]
\item $\textbf{w}$ is the hyperplane normal vector (weight vector)
\item $\nu$ is an upper-bound on the fraction of outliers
and a lower bound on the fraction of support vectors (SV)
\item $\phi$ is a feature mapping function that maps embeddings into a higher-dimensional space (e.g., the radial basis function kernel)
\item $\rho$ is a parameter controlling the margin of the decision boundary
\item $\epsilon_i$ are slack variables allowing for a soft margin
\end{itemize}

To evaluate a generated tuple $t_g$ with embedding $\vec{v}_g$, we project it into the feature space using the kernel function and compute:

\vspace{-4mm}
$$f(\vec{v}_g) = \textbf{w} \cdot \phi(\vec{v}_g) - \rho$$

If $f(\vec{v}_g) \geq 0$, the tuple is deemed acceptable, falling within the normal region defined by the OCSVM. If $f(\vec{v}_g) < 0$, it is potentially deviating from the desired distribution and is rejected.

% By integrating OCSVM into the data augmentation process, we can enhance the quality of generated tuples, ensuring they align with the underlying distribution $\dist$ and mitigating distribution drift.

% Similarly, let $\Sigma_n$ be the covariance matrix of the samples in $\dee$.
% The covariance matrix is computed as: 
% \[
%     \Sigma_n := \frac{1}{n - 1} \sum_{t_i\in\dee}{(\vec{v}_i - \vec{\mu_\dist})}{(\vec{v}_i - \vec{\mu_\dist})^\intercal}
% \]

% We use the inequality proposed by
% Stellato et al.~\cite{stellato2017multivariate}, which only requires the sample mean and covariance matrix to obtain an upper bound on deviation. It computes a threshold $\lambda$ \abol{how is $\lambda$ computed?} and constructs an acceptance region from the
% sample mean and the covariance of the $\dee$. Then, if the Mahalanobis distance of the new sample exceeds $\lambda$, it can be considered as an outlier and we can reject that sample because it is out of the dataset $\dee$ distribution. 

% For a new sample $t$ and its $\kappa$-dimensional vector representation $\vec{v}_t$, if we assume that $\Sigma_n$ is nonsingular, then for all $\lambda \in R_{> 0}$ we have: 

% \[
%     Pr\big((\vec{v}_t - \vec{\mu_\dist})^\intercal \Sigma_n^{-1} (\vec{v}_t - \vec{\mu_\dist}) \ge \lambda^2 \big) \le \frac{\kappa}{\lambda^2}
% \]

% \abol{when exactly do you reject a sample?}

% \abol{please be careful with the notations. size of the dataset is $n$ not $N$. Or did you mean a different thing? I assumed you meant for every tuple in the dataset, is that correct?}.

% \subsection{Challenges}

\subsection{Quality evaluation}\label{sec:test2}
Foundation models have become strong tools for creating high quality multi-modal data.
Still, due to the randomized nature of their generation process, as well as the task-specific difficulties of various queries, some of the generated tuples may not look {\em realistic to human beings}. 

% \abol{the following paragraph is tricky! We should carefully clarify by "quality" as it rightfully can become a point of attack by the reviewers.}

We note that our evaluation of the generated tuples is qualitative and subjective.
That is, the answer to ``does this tuple look realistic'' may vary from one person to the other.
However, if there is good correlation between raters, {\em the probability} of the answer being positive reflects the quality of the tuple.

Using this observation, we model the quality of a tuple as a {\em Bernoulli random variable}.
Specifically, let $p$ be the probability that 
a human evaluator labels a randomly sampled (real) tuple from the distribution $\dist$ as ``realistic''. In other words, with probability $(1-p)$ the evaluator will mistakenly labels the tuple as ``unrealistic''.
We can then define the Bernoulli variable $\phi$, which is one if a real tuple labeled as realistic by a human evaluator, and zero otherwise.
Therefore, the pdf of $\phi$ for the randomly sampled (real) tuples is,
\begin{align}\label{eq:qualitytest}
 f(\phi) = \begin{cases}
  p & \phi=1 \\
  (1-p) & \phi=0
\end{cases}
\end{align}
The mean and the variance of this Bernoulli distribution are $\mu_\phi = p$ and $\sigma^2_\phi = p(1-p)$, respectively. 

Let $p'$ be the probability that a randomly selected human evaluator labels a AI-generated tuple as realistic. When the generated tuple is unrealistic $p'<p$, otherwise the human-evaluator is not better than random labeling. 
We use this observation and develop a hypothesis testing.
Particularly, we discard an AI-generated tuple, if we can reject the null hypothesis that $p'$ is equal to $p$, i.e., $\mathcal{H}_{null}: p' = p$. Then, considering the lower tail test, the alternative hypothesis would be $\mathcal{H}_{alt}: p'<p$.

To do so, we first obtain a sufficiently large sample set $U$ of evaluations, where each sample is drawn using a randomly selected evaluator and a random (real) tuple from $\dee$.
Let $m_U$ be the sample mean of $U$. Since $U$ is sufficiently large, we can estimate $p = \mu_\phi$ with $m_U$. 
%Let $m_U$ and $s_U$ be the sample mean and standard deviation of $U$, respectively.
%Given that $U$ is sufficiently large, we can estimate $\sigma_\phi$ with $s_U$. % Hence, the standard error is estimated as $\frac{s_U}{\sqrt{N}}$.
% Following the central limit theorem, $m_U$ follows the Normal distribution $\mathcal{N}\big(p, \sqrt{\frac{p(1-p)}{N}}\big)$.
Now for a generated tuple $t$, consider a sample set $U_t$ of $N$ evaluations of $t$, each using a randomly selected evaluator. We assume a (small) fixed-size budget for each generated tuple.
Let $m_t$ and $s_t$ be the sample mean and the standard deviation for $U_t$. 
Since $N$ is small, we use the Student's t-test. Specifically,
\[
t_{N-1} = \frac{m_t-p}{s_t/\sqrt{N}}
\]
Next, using the t-table , we obtain the left sided p-value and evaluate its significance level.
If the p-value is smaller than a significance goal $\alpha$, we reject the null hypothesis (discard the generated tuple).

% \subsection{Technical Problems}

%% file: analyzer.tex
\section{Combination Selection}\label{sec:analyzer} 
Our overall goal is to use the foundation model and generate a minimal set of synthetic tuples to resolve the problematic MUPs (with the smallest levels).
Therefore, we consider an iterative approach, where during each iteration we resolve the MUPs at the smallest level.
% At a high level, the goal is to select a small number of combinations that once augmented to the data set, the lack of coverage at the smallest level is resolved.
Given a data set $\dee$, let $\mathcal{M}$ be the set of MUPs, and let $\mathcal{M}^*$ be the set of MUPs with the minimum level. That is $\mathcal{M}^* = \Big\{M\in\mathcal{M} ~|~ \ell(M) = \min_{M'\in\mathcal{M}}\big(\ell(M')\big)\Big\}$.
For each MUP $M\in\mathcal{M}^*$, let us define its gap $\delta(M) = \tau - |\dee\cap M|$; i.e., the coverage threshold minus the current coverage of $M$ in $\dee$. In other words, $\delta(M)$ is the minimum number of synthetic tuples matching $M$ we need to obtain before it is covered.
Also, for each combination $c_i\in \varprod_{k=1}^d dom(x_k)$, let $\sigma_i$ be the number of synthetic tuples from that combination.
Then, the ``\cs'' problem is to 
assign the values of $\sigma_i>0$ such that (i) for each MUP $M\in\mathcal{M}^*$, at least $\delta(M)$ generated tuples match it, and (ii) sum of all $\sigma_i$ values is minimized. Formally,

\vspace{-5mm}
\begin{align*}
    \min &\hspace{6mm}\sum_{c_i} \sigma_i\\
    \mbox {Subject to} &\hspace{2mm}\underset{c_i\in \mbox{match}(M)}{\sum \sigma_i} \geq \delta(M), \quad\forall M\in\mathcal{M}^*
\end{align*}

\begin{lemma}\label{lem:nphardness}
    \cs is \np-hard.\footnote{Proofs are provided in \S~\ref{sec:proofs}.}
\end{lemma}

\begin{algorithm}[!tb]
    \caption{\greedy} \label{alg:greedy}
    \begin{algorithmic}[1] \small
    \Require{The smallest-level MUPs $\mathcal{M}^*$}
    \Ensure{The number of instances from each combination to be augmented to $\dee$}
        \State $\sigma\gets$ {\em new} {\bf hashmap}$()$
        \For{$M\in\mathcal{M}^*$}
           \State $\delta(M)\gets \tau - |\dee\cap M|$
        \EndFor
        \While{$\mathcal{M}^*$ is not empty}
            \State find the combination $c$ that matches most MUPs in $\mathcal{M}^*$
            \State $\mathcal{M}'\gets\{M\in\mathcal{M}^*~|~c \mbox{ matches } M\}$
            \State $\gamma \gets \min_{M\in\mathcal{M}'}\delta(M)$
            \If{$c\in \sigma.keys$}
                \State $\sigma[c]\gets \sigma[c]+\gamma$
            \Else
                $\quad \sigma[c]\gets \gamma$
            \EndIf
            \For{$M\in\mathcal{M}'$}
                \State $\delta(M)\gets \delta(M)-\gamma$
                \If {$\delta(M)=0$}
                    \State Remove $M$ from $\mathcal{M}^*$
                \EndIf
            \EndFor
        \EndWhile
        \State {\bf return} $\sigma$
    \end{algorithmic}
\end{algorithm}

Since \cs is \np-hard, we design an approximation algorithm for this step. Our algorithm follows the {\em greedy} scheme. 
Algorithm~\ref{alg:greedy} shows the pseudo-code of the \greedy algorithm.
The algorithm is iterative, where at each iteration it finds the combination that matches the maximum number of remaining MUPs in $\mathcal{M}^*$. We utilize the inverted index and the tree data structure proposed in \cite{asudeh2019assessing} for finding $c$.
The algorithm then finds the minimum gap $\gamma$ in the MUPs matching $c$ and increases the number of instances from $c$ by $\gamma$.
It also updates the gaps for the MUPs matching $c$
 and remove the ones that reach to a gap of zero from $\mathcal{M}^*$.

\begin{theorem}\label{th:greedy}
    The approximation ratio of the \greedy approach is $\log(\eta)$, where $\eta=\sum_{M\in\mathcal{M}^*}\delta(M)$.
\end{theorem}

\begin{comment}
In this section, we elaborate on the process of identifying insufficient coverage within a given dataset $\dee$, guided by the definition of coverage as outlined in \S~\ref{sec:pre:datacoverage}. Our approach involves identifying Maximal Uncovered Patterns (MUPs) using the method proposed in \cite{asudeh2019assessing}. For a comprehensive understanding, we provide an explanation of the underlying mechanics of this algorithm by \cite{asudeh2019assessing}.
\mahdi{TODO: Expand on how this algorithm works}

Once the MUPs are identified, our objective is to determine the optimal combinations for repair. We focus on identifying children of the identified MUPs and subsequently selecting the most effective combination for repair.

To address the task of selecting the repair combination, we frame the problem as a hitting set problem. The Set of identified MUPs is denoted as $P = \{P_1, P_2, \dots, P_k\}$, where each MUP $P_i$ may have multiple children. The number of children for each MUP is derived from its level and the cardinality of unknown attributes. For a MUP $P_i$ with unknown attributes of interests $u = \{u_1, u_2, \dots, u_j \}$ where $u \subseteq x$, it has $p = \prod_{i=1}^{j} dom(u_i) $ children.

We propose a greedy algorithm leveraging this graph structure to target the leaf nodes with the maximum number of parents.

\mahdi{TODO: Write details of greedy algorithm }
\end{comment}

%% file: selection.tex
\section{Guide tuple Selection}\label{sec:selection}
% \subsection{Introduction}

Given a combination $c$, we would like to generate a tuple that matches $c$ and is likely to pass the rejection sampling tests.
Therefore, we want to make sure that (a) the generated tuple follows the underlying distribution $\dist$ represented by $\dee$ and (b) the generated tuple has a high quality and passes the quality evaluation.
So, instead of relying on the foundation models imagination, we provide a ``guide'' for the generation process.
% the goal of the guide tuple selector is to sample an tuple from the data set $\dee$, to be used as the guide for generating the next tuple.
% \subsection{Problem Formulation}
% Given a combination $c$, the first task is to select a guide tuple $t$ from the data set $\dee$ that serves as a guide for generating a tuple matching $c$. 
Recall from \S~\ref{sec:pre:foundationmodel}, that the guide is a pair $(t,m)$, where $t$ is a tuple and $m$ is a mask.
In the following, we propose various strategies for guide tuple selection. We begin with a baseline approach discussed in \S~\ref{sec:selection:random} and progressively explore more sophisticated strategies in  \S~\ref{sec:selection:similar} and \S~\ref{sec:selection:ucb}. 

In the context of images, a mask specifies the parts of the guide image that should be regenerated. Our mask generation (\S~\ref{sec:selection:mask}) 
% In the context of images, next steps 
involve cropping the foreground of $t$ using a mask $m$, ensuring that the foundation model regenerates only the portions delineated by the mask $m$. 
% The details of mask delineation process are discussed in \S~\ref{sec:selection:mask}.

\subsection{Random-Guide Strategy}\label{sec:selection:random}
The random guide strategy 
focuses on the first requirement that: the generated tuple should follow the underlying distribution $\dist$, represented by $\dee$. Then, viewing each tuple $t$ as a random sample from $\dee$, it selects the guide tuple uniformly at random from the data set without taking into account the target combination $c$.
While the random-guide strategy is appropriate for satisfying the underlying distribution, it ignores the second requirement of passing the quality test. 
% So, generated tuples may combine features one expects to see in different parts of the feaible space, and hence appear unrealistic. 
We  experimentally show in \S~\ref{sec:exp}, that tuples generated based on this strategy have a lower chance of passing quality evaluation.

% The random strategy involves selecting a guide tuple uniformly at random from the data set $\dee$ without taking into account the target combination $c$. While this strategy may result in the selection of an inappropriate tuple (e.g.  choosing an infant tuple as the guide for generating an elderly one), it mitigates the risk of repeatedly choosing the same tuple. We employ this strategy as a baseline in our experiments, as detailed in \S~\ref{sec:exp}.

% \mahdi{I couldn't tell it very well but if the attributes that we have in our data set are not so different from each other that make tuple switching easier, then we can use this strategy, for example if we only had gender and race in our data set, then this strategy might prodcue acceptable resutls but when we have age attribute, tuples of each age group is very different from other age group, so we need  to be in similar zone for base tuple to get acceptable results}

\subsection{Similar-Tuple Strategy}\label{sec:selection:similar}
The similar-tuple strategy creates a pool of {similar} combinations to the target combination $c$. 
% moved the definition of "combination" to the end of 2.1.
% To assess the connection between two combinations, we define the notion of {\em similarity}. 
Combinations $c_1$ and $c_2$ are considered {\em similar} if (a) they are {\em siblings} (i.e., their values differ in exactly one attribute) and (b) one of the following conditions is satisfied.
Let $d_i$ be the attribute on which $c_1$ and $c_2$ differ.
If $d_i$ is non-ordinal then $c_1$ and $c_2$ are considered similar.
However, if $d_i$ is ordinal, then the distance between $c_1$ and $c_2$ should be 1 to be considered similar.
Formally, for two sibling combinations $c_1$ and $c_2$ that differ in attribute $d_i$:

\vspace{-7mm}
\[
\mbox{similar}(c_1, c_2) = 
\begin{cases} 
\text{\em false} & \text{if } d_i \text{ is ordinal and } |c_1[d_i] - c_2[d_i]| > 1\\
\text{\em true} & \text{otherwise}
\end{cases}
\]

\begin{comment}
under two conditions. Firstly, they must differ in only one digit, denoted as $d_i$. Secondly, if $d_i$ corresponds to an ordinal attribute, then $|c_1[d_i] - c_2[d_i]| = 1$. This constraint ensures that when dealing with an ordinal attribute, only adjacent categories are considered for comparison. 

\[
\mbox{similar}(c_1, c_2) = 
\begin{cases} 
true & \text{if } \exists ~d_i \text{ s.t. } \begin{cases} 
    c_1[d_i] \neq c_2[d_i] \\
    \text{if } d_i \text{ is ordinal, then } \\ \text{ \space} |c_1[d_i] - c_2[d_i]| = 1 \\
    \forall d_j \neq d_i, c_1[d_j] = c_2[d_j]
\end{cases} \\
false & \text{otherwise}
\end{cases}
\]

\end{comment}

Subsequently, it selects a guide tuple from the pool of similar combinations, assigning weights to each element based on the number of tuples in $\dee$ that adhere to that particular combination. Formally, the pool of similar combinations can be defined as follows:
\[
S = \Big\{ c \in \mbox{\em sibling}(c_i) ~\Big|~ \mbox{similar}(c_i, c) = \text{\em true} \Big\} \\ 
\]

% \[
% S = \Big\{ c \in \varprod_{k=1}^d dom(x_k) ~\Big|~ \mbox{similar}(c_i, c) = true \Big\} \\ 
% \]
% In this definition, $C = \varprod_{k=1}^d dom(x_k)$ represents the set of all possible combinations, and $\text{similarity}(c_i, c)$ denotes the previously defined similarity function between combinations $c_i$ and $c$.
For each combination $c_i\in S$, let $|c_i|$ be the number of tuples in $\dee$ matching it. That is, $|c_i|= |c_i\cap \dee|$.
We assign the sampling weight of each combination proportional to their normalized size:
% Moreover, the formal expression for the weight of $s_i \in S$ is given by: 
\[
w_i = \frac{|c_i|}{\sum_{c_j \in S} |c_j|}, ~~ \forall c_i\in S
\]
% The weights $W_i$ are computed based on the size of each subset $s_i$ in $S$ relative to the total size of all subsets in $S$, providing a normalized measure of importance for each combination in the pool.

The similar-tuple strategy then selects a combinations $c_i\in S$, randomly with probability $w_i$. It then returns random sample from the pool tuples in $\dee$ that match $c_i$, as the guide tuple. Using $w_i$ as the weight ensures equal sampling probability for all tuples that match a combination in $S$.

This strategy considers tuples in the selection pool that closely resemble the target combination, differing in only one attribute of interest. It excludes tuples that are far away from the target combination. It also excludes the tuples  with the exact combination as the target combination.
This exclusion is intentional to deal with the fact that the target combination $c$ is not very common in the data set. Since $c$ is not well-represented, picking guide tuples from this group might make all the chosen tuples look too similar. By considering combinations that are similar but not exactly the same as the target one, the strategy aims to make sure we get a more varied and representative set of guide tuples for the generation process.

\subsection{Modeling as Contextual Multi-armed  Bandit}\label{sec:selection:ucb}
Our \linucb strategy models
% In this section, we model 
the guide tuple selection problem as a {\em contextual multi-armed bandit} problem \cite{bouneffouf2020survey}, and uses  {\em Contextual Upper Confidence Bound} for solving it~\cite{li2010contextual}.
Specifically, it {models each attribute as a bandit arm}. Then
% Our \linucb \cite{li2010contextual} strategy is an attribute-aware approach designed 
given %\nima{given?} 
a target combination $c$, it selects an arm to pull (i.e., an attribute to modify), aiming to maximize the obtained {reward}. In each iteration, we have the opportunity to pull only one arm, signifying the ability to alter one attribute value within the target combination to a new value. The reward obtained from pulling that arm is then observed. The objective is to learn the optimal arm to pull for a given combination $c$ over successive iterations. 

To provide further clarification, let us %\nima{to short forms in formal writing: let us} 
discuss an example within the context of images. Consider a data set with attributes of interest including \at{gender}, \at{race}, and \at{age-group}. The foundation model \fm may perform better in modifying the race of a subject compared to altering their age group for specific combinations (e.g., Asian female adults). However, its performance may vary for other combinations. \linucb aims to systematically explore different arms to pull (e.g., changing race, gender, or age group) and exploit the arm yielding the highest reward over time. 

Formally, we formulate the guide tuple selection problem as a contextual multi-armed bandit problem. We consider the attributes of interest, denoted as $\mathbf{x} = \{ x_1, x_2, \dots, x_d \}$, as arms of the bandit $\mathbf{a} = \{ a_1, a_2, \dots, a_d \}$.
% each arm $a_i$ reward behavior could be modeled as a real distribution $\dist_i$ with associated mean of $\mu_i$ in given context $\mathbf{c}$. 
The {\em context}, AKA the {\em feature vector}, is then defined as a one-hot vector $\textbf{f}$ representing combinations, where 1 is assigned for the input combination $c$ and 0 for all other elements. Let $k = |\varprod_{i=1}^{d}dom(x_i)|$ be the number of possible combinations. The size of the feature vector $\textbf{f}_{s,a}$ is $(k \times 1)$  where $s$ denotes the time step of the algorithm. 

% \abol{Mahdi, please check the following, make sure it is correct}

We define the reward function based on whether a generated tuple passes the rejection sampling tests. Let {\em pass$()$} be a binary function that is false if a generated tuple is rejected. Then, 

\vspace{-4mm}
\[
r_{s,a} = 
\begin{cases} 
1 & \text{if } \text{\em pass}()= \text{\em true} \\
0 & \text{otherwise}
\end{cases}
\]

We adopt  ``LinUCB with Disjoint Linear Models''~\cite{li2010contextual} to balance exploration and exploitation. At every iteration $s$, for
every arm $a \in \textbf{a}$, given the context $\textbf{f}_{s,a}$, \linucb computes confidence intervals for the expected reward and selects the arm with the maximum
upper bound of reward to be explored next. 

We assume that the expected reward of an arm $a$ is linear in its $k$-dimensional input context $\textbf{f}_{s,a}$ with some unknown coefficient vector $\theta^*_{s,a}$, formally given by: 

\vspace{-7mm}
\[
    \hspace{30mm}E[r_{s,a}| \textbf{f}_{s, a}] = \textbf{f}_{s,a}^\top \theta_{s,a}^*
\]
Assuming that $m$ is the number of times arm $a$ has been pulled so far ($s \ge m)$, we define $\textbf{F}_{s,a} $ with size $(m \times k)$ as the matrix consisting of all previously observed contexts for arm $a$. 
% \[
%     \textbf{F}_{s, a} =  \begin{bmatrix}
%                                 \textbf{f}_{1,a}^\top \\
%                                 \vdots \\ 
%                                 \textbf{f}_{m,a}^\top \\
%                         \end{bmatrix}
% \]
\[
    \textbf{F}_{s, a} =  \big[
                                \textbf{f}_{1,a}^\top,
                                \cdots,
                                \textbf{f}_{m,a}^\top
                        \big]^\top
\]
We also have a vector of observed rewards from pulling arm $a$. 
% \[
%     \Gamma_{a} =  \begin{bmatrix}
%                                 {r}_{1,a} \\
%                                 \vdots \\ 
%                                 {r}_{m,a} \\
%                         \end{bmatrix}
% \]
\[
    \Gamma_{a} =  \big[
                                {r}_{1,a},
                                \cdots,
                                {r}_{m,a}
                        \big]^\top
\]
Vector $\textbf{b}_{a}$ is defined as:

\vspace{-7mm}
\[
    \textbf{b}_{a} = \textbf{F}_{s, a}^\top \Gamma_{a}
\]
Using Ridge regression estimator, we can estimate the coefficients of each arm $a$ as: 

\vspace{-6mm}
\[
    \hspace{20mm}\hat{\theta}_{s, a} = (\textbf{F}_{s, a}^\top \textbf{F}_{s, a} + \textbf{I}_k)^{-1} \textbf{b}_{a}
\]
Thus, in each iteration $s$ of the algorithm, we select arm $a_s$ using:
\[
    a_s = \argmax_{a \in \textbf{a}} \big( \textbf{f}_{s,a}^\top \hat{\theta}_{s,a} + \alpha\sqrt{\textbf{f}_{s,a}^\top \textbf{A}_{a}^{-1} \textbf{f}_{s,a}}\big)
\]
where $\textbf{A}_{a} = \textbf{F}_{s, a}^\top \textbf{F}_{s, a} + \textbf{I}_k$ and $\alpha$ is a hyper-parameter to balance exploitation and exploration. 

After pulling arm $a_s$ in iteration $s$, we observe the reward $r_{s,a_s} \in \{0, 1\}$ where 1 indicates that the generated tuple $t$ has passed quality and data distribution tests. We can update matrices $\textbf{A}_{a_s}$ and $\textbf{b}_{a_s}$ as: 

\vspace{-2mm}
\[
    \textbf{A}_{a_s} \leftarrow \textbf{A}_{a_s} + \textbf{f}_{s, a_s} \textbf{f}_{s, a_s}^\top \\
\]
\[
    \textbf{b}_{a_s} \leftarrow \textbf{b}_{a_s} + r_{s,a_s} \textbf{f}_{s, a_s}
\]

% After 

Algorithm \ref{alg:linucb} presents the pseudo-code for the \linucb strategy used in selection of guide tuple $t$ from the dataset $\dee$. 

\begin{algorithm}[!tb]
    \caption{\linucb} \label{alg:linucb}
    \begin{algorithmic}[1] \small
        \Require{Attributes of interest as arms $\mathbf{a} = \{ a_1, \dots, a_d \}$, exploration parameter $\alpha$, $k = |\varprod_{i=1}^{d} dom(x_i)|$}
        \Ensure{Sequence of selected guide tuples and observed rewards}

        \For{$a \in \textbf{a}$}
            \State $\mathbf{A}_a \leftarrow \mathbf{I}_k$, $\mathbf{b}_a \leftarrow \mathbf{0}_{k\times 1}$ \Comment{Initialize matrices}
        \EndFor
        
        \For{each iteration $s$}
            \State $c_s \leftarrow$ {\bf Input} $()$ \Comment{Get target combination}
        \For{$a \in \mathbf{a}$}
            \State $\mathbf{f}_{s,a} \leftarrow $ {\bf Convert} $(c_s)$ \Comment{Calculate context vector from $c_s$}
            
            \State $\hat{\theta}_{s,a} = \mathbf{A}_{a}^{-1} \mathbf{b}_a$ \Comment{Estimate coefficients}
            
            \State $UCB_{s,a} = \mathbf{f}_{s,a}^\top \hat{\theta}_{s,a} + \alpha\sqrt{\mathbf{f}_{s,a}^\top \mathbf{A}_{a}^{-1} \mathbf{f}_{s,a}}$ \Comment{Calculate UCB}
        \EndFor

        \State $a_s = \argmax_{a \in \mathbf{a}} UCB_{s,a}$ \Comment{Select arm $a_s$}

        \State $c' \leftarrow $ {\bf Modify} $(c, a_s)$ \Comment{Modify attribute $a_s$ of $c$ to create new combination}
        
        \State $t \leftarrow$  {\bf Select}$(c')$ \Comment{Select tuple $t$ matching $c'$}

        \State $r_{s,a_s} \leftarrow $ {\bf Input}$()$ \Comment{Observe reward for $r_{s,a_s}$}

        \State $\mathbf{A}_{a_s} \leftarrow \mathbf{A}_{a_s} + \mathbf{f}_{s,a_s} \mathbf{f}_{s,a_s}^\top$ \Comment{Update matrices}
        
        \State $\mathbf{b}_{a_s} \leftarrow \mathbf{b}_{a_s} + r_{s,a_s} \mathbf{f}_{s,a_s}$

        \EndFor

    \end{algorithmic}
\end{algorithm}

\subsection{Mask Delineation}\label{sec:selection:mask}

\begin{figure}[!tbp]
\begin{tcolorbox}[
    % enhanced,
    colback=gray!10,
    colframe=gray!10,
    boxrule=0.5pt,
    rounded corners,
    boxsep=2pt,
    width=0.48\textwidth
]
  \centering
  \begin{subfigure}{0.23\textwidth}
    \includegraphics[width=\linewidth]{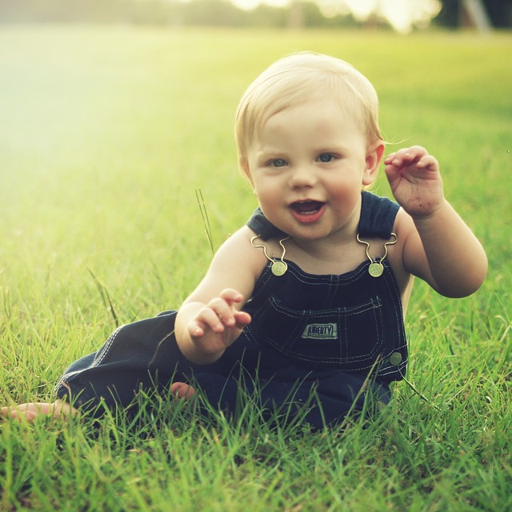}
    \caption{}
  \end{subfigure}
  \hfill
  \begin{subfigure}{0.23\textwidth}
    \includegraphics[width=\linewidth]{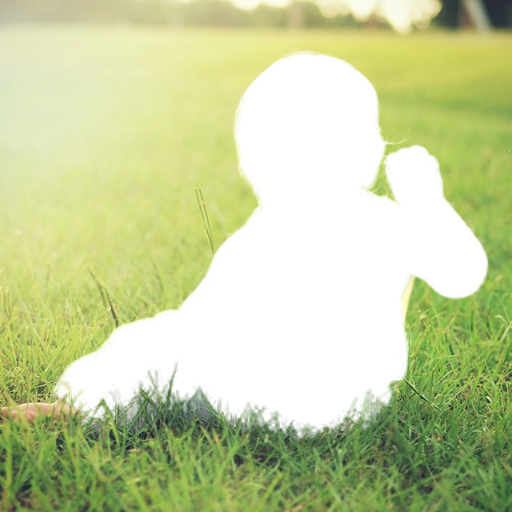}
    \caption{}
  \end{subfigure}
  \hfill
  \begin{subfigure}{0.23\textwidth}
    \includegraphics[width=\linewidth]{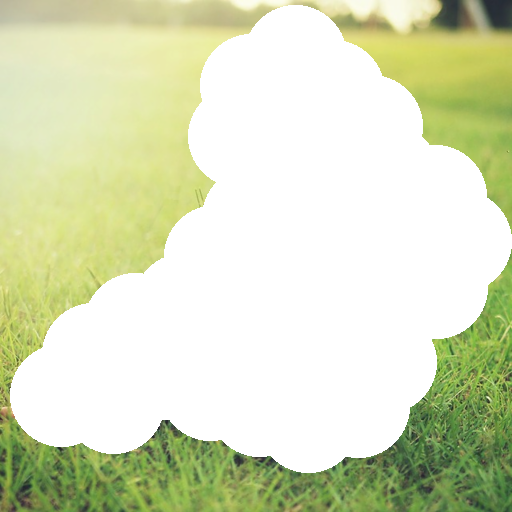}
    \caption{}
  \end{subfigure}
  \hfill
  \begin{subfigure}{0.23\textwidth}
    \includegraphics[width=\linewidth]{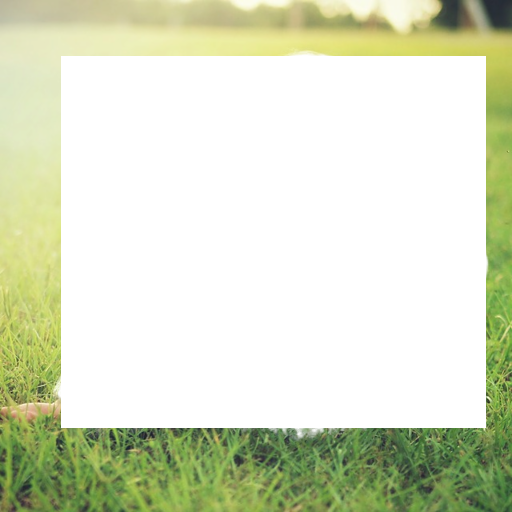}
    \caption{}
  \end{subfigure}
\vspace{-4mm}
  \caption{Illustration of various guide image (a) masks: (b) Accurate (c) Moderate (d) Imprecise mask}
  \label{fig:mask:comparison}
\end{tcolorbox}
\vspace{-6mm}
\end{figure}

In the context of images, once the guide tuple $t$ is selected, we delineate the foreground subject using a mask. This mask serves as an indicator, specifying the regions to be cropped and regenerated from the tuple $t$. The delineation of the border around the subject can be achieved with different levels of precision. A precise border sketch preserves more space from the original context, potentially resulting in a higher acceptance rate for data distribution test. However, it may limit the foundation model imagination capacity and lead to lower acceptance rate for quality evaluation test. We propose three levels of mask sketch accuracy: {\em accurate}, {\em moderate}, and {\em imprecise} (Figure~\ref{fig:mask:comparison}).

\subsubsection{Accurate mask delineation}\label{sec:selection:mask:acc}
This represents the highest level of precision in mask delineation, achieved by utilizing the off-the-shelf background remover tool \at{rembg}.
\subsubsection{Moderate mask delineation}
To obtain a moderately delineated mask, we extend the border of the mask drawn in \ref{sec:selection:mask:acc} by 10 percent of the image size. This extension is implemented using circles, with each point on the mask border being surrounded by a circle of radius equal to $10\%$ of the image width.
\subsubsection{Imprecise mask delineation}
For an imprecise mask delineation, we expand the border of the mask drawn in \ref{sec:selection:mask:acc} to form a rectangular area. This rectangle encompasses the previous mask, providing a less precise but more inclusive delineation.

% Figure \ref{fig:mask:comparison} illustrates the impact of different mask delineation levels on the guide image. 

%% file: exp.tex
\section{Experiments}\label{sec:exp}

%\subsection{Experiments Plan}\label{sec:exp:plan}

This section presents results from experiments to evaluate the efficacy of our proposed system, \system. We begin by providing a concise overview of its implementation details and noteworthy technical challenges encountered during development (\S~\ref{sec:exp:imp}). Next, we delve into the experimental setup, outlining the datasets, monetary cost considerations, baselines, and performance metrics used in the evaluation (\S~\ref{sec:exp:setup}).
We then present a proof-of-concept demonstration for the paper's core proposition: {\em the efficient and effective resolution of lack of coverage for under-represented groups} (\S~\ref{sec:exp:poc}). This demonstration serves as a foundational validation of the system's functionality.

Finally, we conduct a comprehensive performance evaluation of individual design decisions and system components across three distinct tasks. Each task leverages a specific benchmark, and we present comparative analyses against state-of-the-art baselines to objectively assess the system's effectiveness.

\begin{enumerate}[label=\numberingI{\arabic*},leftmargin=*]
    % \item We perform a detailed analysis of the associated costs with repairing Maximal Uncovered Patterns (MUPs) at different levels. Additionally, we explore the relationship between the number of resolved MUPs and the corresponding costs incurred by \fm. \label{sec:exp:tasks:mupsandcost}
    
    \item We investigate the system's performance in passing the data distribution test (\S~\ref{sec:test1}) and quality evaluation test (\S~\ref{sec:test2}), using various mask delineation levels and guide tuple strategies. \label{sec:exp:tasks:dataquality}
    
    \item We analyze the performance of the proposed \greedy approach (Algorithm~\ref{alg:greedy}) for combinations selection. \label{sec:exp:tasks:combselection} 

    \item %Given our agnostic stance on the choice of quality assessment tools, whether human evaluators or automated tools, we explore alternative options to replace human evaluators. 
    Our quality evaluation test (\S~\ref{sec:test2}) involves human evaluators.
    In presence of automated tools that perform similar to human evaluators, this step could be automated. In this experiment, we explore alternative options to replace human evaluators.
    We analyze the results obtained from different quality assessment tools and compare them to the ground truth by human evaluators. \label{sec:exp:tasks:humanevaluator}
    
    % \item  We illustrate the consequences of lack of coverage on a race detection model using an unfair data set. Subsequently, we employ \system to enhance coverage and augment data set using synthetic data and report new accuracy and F1 scores for the model. \label{sec:exp:tasks:ml}
\end{enumerate}

\subsection{Implementation Details}\label{sec:exp:imp}
% In this section, we explore the technical complexities of the \system implementation, briefly addressing the encountered technical challenges.

\paragraph{Architectural style} The implementation of \system is based on a microservices architecture, providing modularity and independence for each project component. Employing microservices allows us to develop each aspect of the project as an independent service, communicating through {\sc {APIs}}. It also gives us the ability to replace each service without interrupting others if required. All backend microservices are implemented using the {\sc {FastAPI}} framework, while the frontend service is developed using {\sc {React JS}}. The various services in our project include Gateway, ImageAnalyzer, ImageEditor, MaskGenerator, PreProcessor, LinUCB, and Chameleon-UI.

\paragraph{Integration with Foundation Model}
For seamless integration with \dalle, specific formatting of input and output image sizes is imperative. All input images are required to be converted into square images with dimensions of $256x256$, $512x512$, or $1024x1024$. To meet this criterion, we employ two distinct strategies.
% \begin{itemize}
    % \item 
    {\em 1) Cropping:} 
    when dealing with data sets consisting images characterized by varying dimensions (portrait and landscape), we filter out images with a ratio of the larger side to the smaller side exceeding 1.5. This filtering ensures the keeping of predominantly square images in our data set. Subsequently, we perform a center crop and resize of all images to the nearest acceptable square size.
    % \item 
    {\em 2) Encasing:}
    in instances where all images share uniform dimensions and exhibit elongated rectangular shapes, direct cropping may result in the loss of important features. To address this, we encase each image within a larger square image with a white background. Post \dalle image generation, we transform the resulting tuple to match the dimensions of the original data set, effectively removing the surplus white area.
% \end{itemize}

\subsection{Experimental Setup}\label{sec:exp:setup}

\subsubsection{Hardware Configurations}
The majority of the development and testing for \system occurred on a personal computer equipped with 12 {\sc x86} cores and 16 gigabytes of memory. For experiments that required training a CNN model, we utilized an {\sc Ubuntu} Linux server featuring 24 {\sc x86} cores and 128 gigabytes of memory.
\subsubsection{Data Sets}

\begin{figure}
\begin{tcolorbox}[
    % enhanced,
    colback=gray!10,
    colframe=gray!10,
    boxrule=0.5pt,
    rounded corners,
    boxsep=2pt,
    width=0.48\textwidth
]
  \centering

  \begin{subfigure}{0.23\textwidth}
    \includegraphics[width=\linewidth]{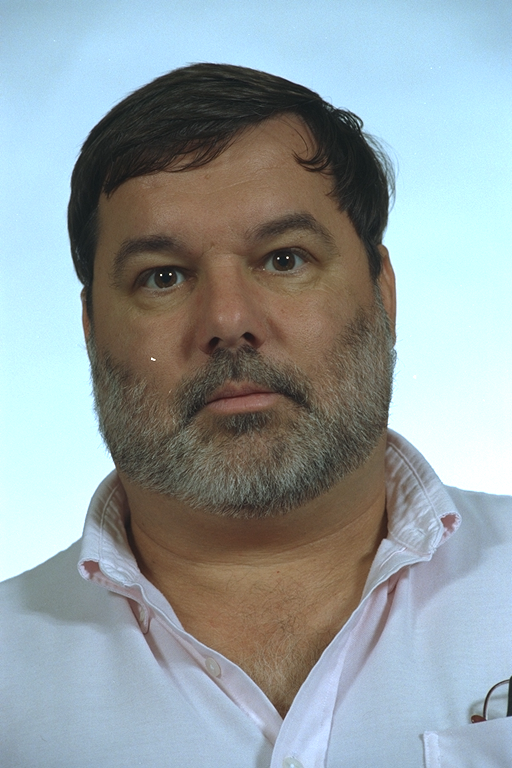}
    \caption{}
  \end{subfigure}
  \hfill
  \begin{subfigure}{0.23\textwidth}
    \includegraphics[width=\linewidth]{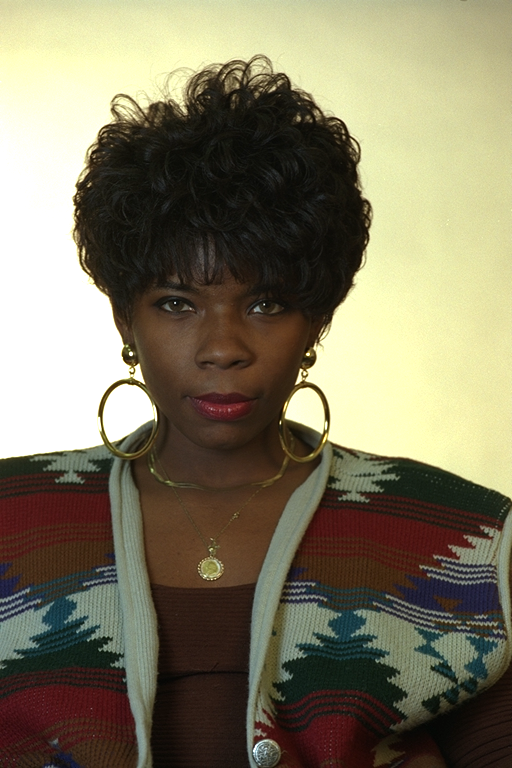}
    \caption{}
  \end{subfigure}
  \hfill
  \begin{subfigure}{0.23\textwidth}
    \includegraphics[width=\linewidth]{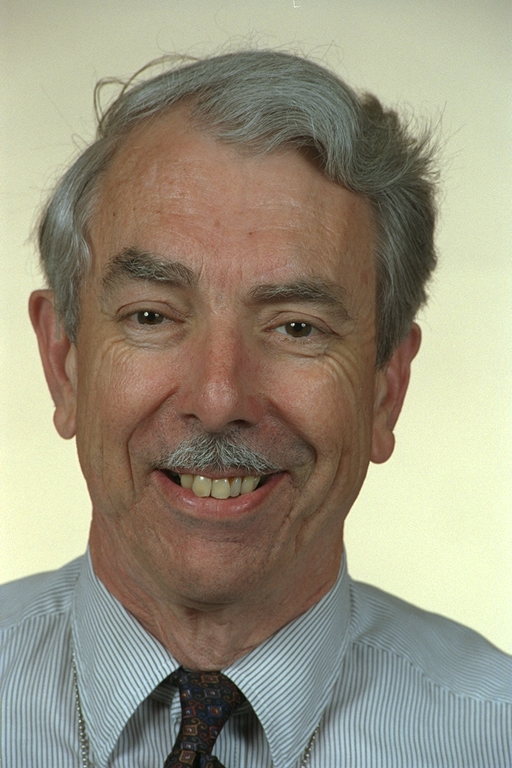}
    \caption{}
  \end{subfigure}
  \hfill
  \begin{subfigure}{0.23\textwidth}
    \includegraphics[width=\linewidth]{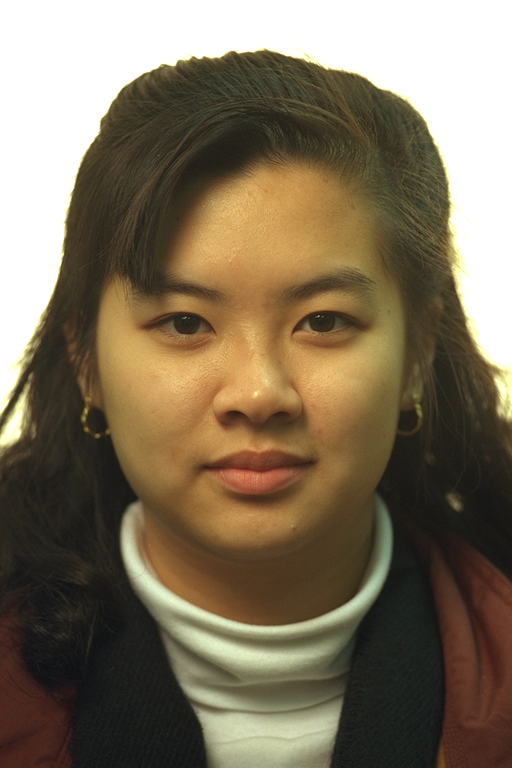}
    \caption{}
  \end{subfigure}

  \medskip

  \begin{subfigure}{0.23\textwidth}
    \includegraphics[width=\linewidth]{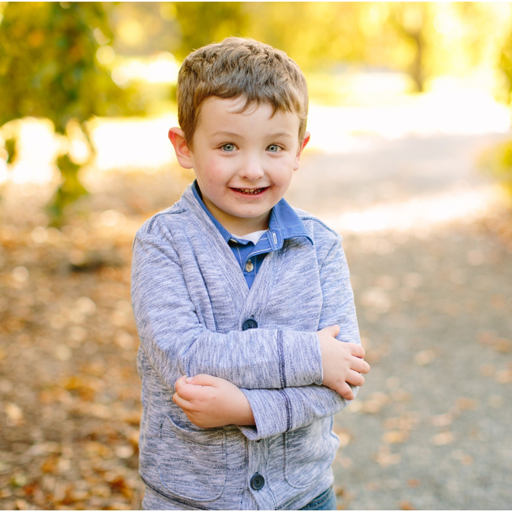}
    \caption{}
  \end{subfigure}
  \hfill
  \begin{subfigure}{0.23\textwidth}
    \includegraphics[width=\linewidth]{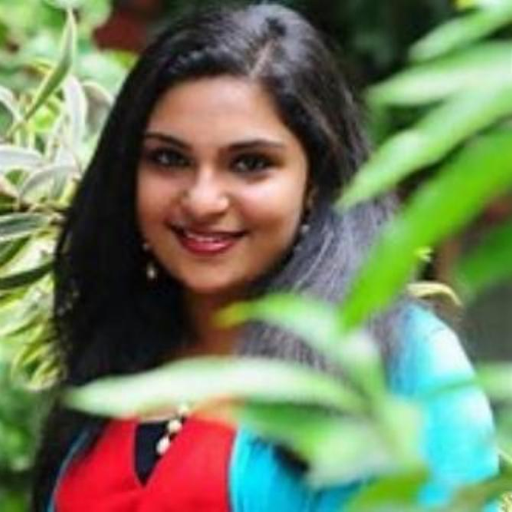}
    \caption{}
  \end{subfigure}
  \hfill
  \begin{subfigure}{0.23\textwidth}
    \includegraphics[width=\linewidth]{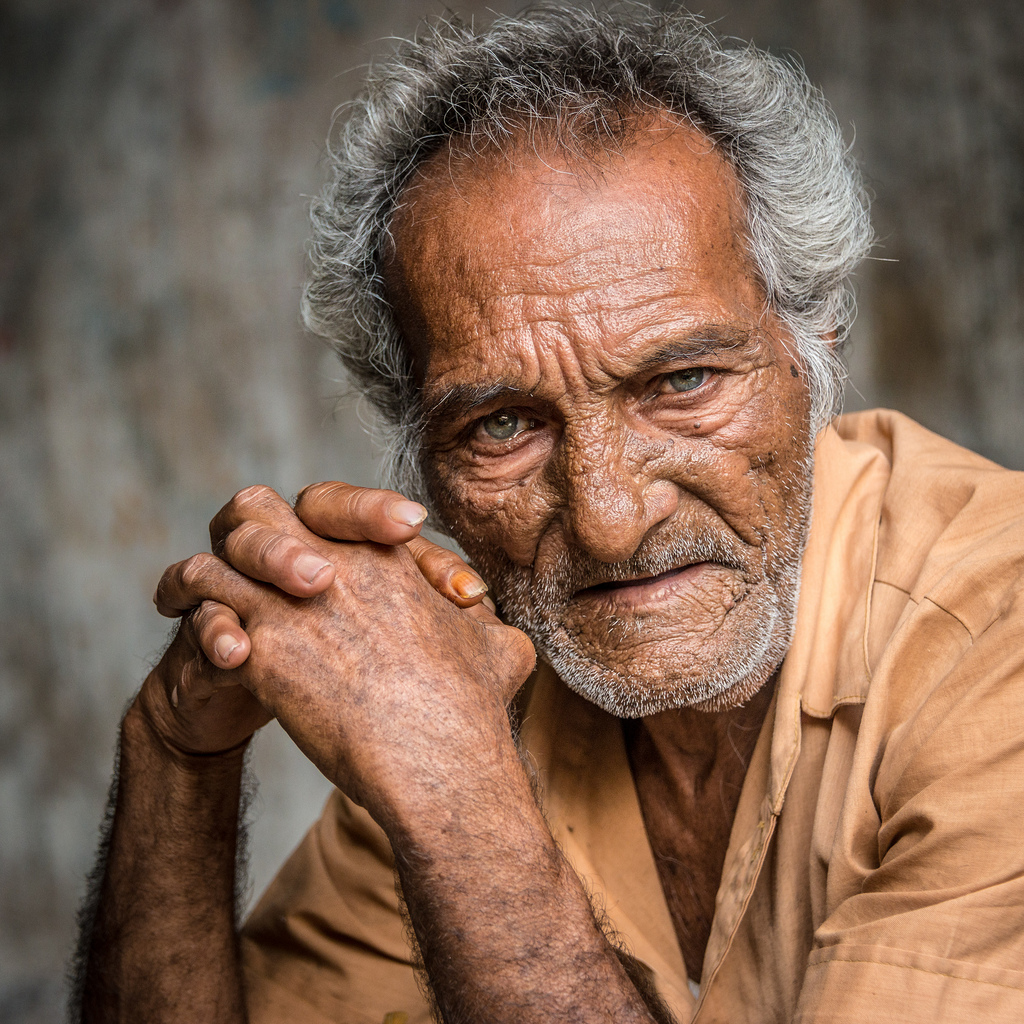}
    \caption{}
  \end{subfigure}
  \hfill
  \begin{subfigure}{0.23\textwidth}
    \includegraphics[width=\linewidth]{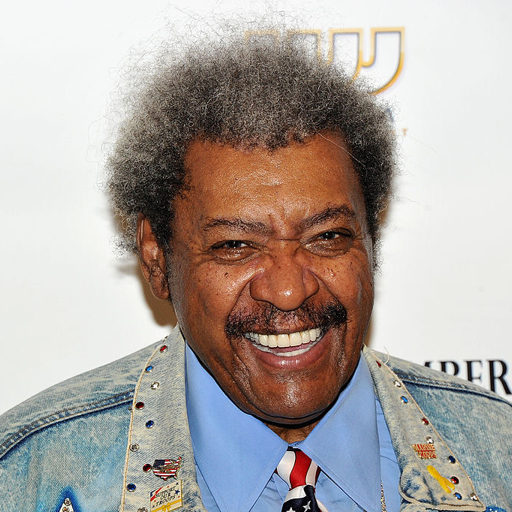}
    \caption{}
  \end{subfigure}
\vspace{-3mm}
  \caption{Sample images from \feretdb  (a-d) and \utkface (e-h) data sets}
  \label{fig:multiimages}
\end{tcolorbox}
\vspace{-5mm}
\end{figure}

We utilize distinct subsets from \utkface \cite{zhifei2017cvpr} and \feretdb \cite{phillips1998feret} in our experiments. \utkface encompasses over 20,000 face images with annotation of age, gender, and ethnicity. The images in \utkface cover large variation in pose, facial expression, illumination, occlusion, resolution, and other factors. Conversely, \feretdb comprises 1199 individual images and serves as a standardized facial image database for researchers to develop algorithms and report results. All images in \feretdb share the same dimensions, pose, and facial expression and are annotated with gender and ethnicity. Figure~\ref{fig:multiimages} presents actual samples from both data sets.

\subsubsection{Foundation Model and the Monetary Cost}
We use \dalle Foundation model for generating images from prompt as it is (at the time of experiments) the most widely available Image Generation model with public {\sc API KEY} available. Throughout the development and experimental phases, a total of 1962 distinct images were generated using \dalle. {\em The total cost for generating these images amounted to (US)\$33.6}. 

\subsubsection{Evaluated Algorithms and Baselines}
The following are the baselines and evaluated algorithms designed for each task.
\begin{itemize}[leftmargin=*]
    \item 
    For Task~\ref{sec:exp:tasks:dataquality}, we consider {\sc no-guide tuple}, along with {\sc similar-tuple} and {\sc random-guide} strategies, as baselines for our experiments to compare against \linucb. Additionally, we consider {\sc accurate}, {\sc moderate}, and {\sc imprecise} mask delineation levels for evaluation. 
    \item 
    For Task~\ref{sec:exp:tasks:combselection}, we consider two baselines to compare against the \greedy combination selection algorithm. The first baseline, called {\sc random}, randomly selects the next combination to be generated. The second baseline ({\sc Min-Gap}), however, first identifies a MUP $M$ that requires the minimum number of instances to be covered. It then generates a combination that matches $M$. Both {\sc random} and {\sc Min-Gap} baselines continue until the MUPs at smallest level are resolved.
% compare the results of the \greedy combination selection algorithm with those of the {\sc random} combination selection strategy and the {\sc Min-Gap} selection strategy. 
    \item 
    For Task~\ref{sec:exp:tasks:humanevaluator}, we employ state-of-the-art image quality assessment tools, {\sc NIMA} \cite{talebi2018nima}, {\sc BRISQUE} \cite{mittal2012no} and {\sc NIQE} \cite{mittal2012making} as baselines for comparing results with human ground truth. 
    \item 
    Lastly, for our proof of concept, we designate the {\sc Tensorflow Keras} CNN Model for precision, recall and F1 score comparisons.
\end{itemize}
% \begin{enumerate}[label=\numberingI{\arabic*}]
%     \item[\ref{sec:exp:tasks:combselection}] Combination selection:  random combination, greedy algorithm 
%     \item[\ref{sec:exp:tasks:dataquality}]  Guide image selection: no guide image - random guide image - similar - ucb
%     \item[\ref{sec:exp:tasks:dataquality}] Mask sketch: accurate, imprecise, moderate 
%     \item[\ref{sec:exp:tasks:humanevaluator}] image assessment tools: NIMA, NIQE and BRISQUE (baselines), Ground truth from human evaluators 
%     \item[\ref{sec:exp:tasks:ml}] CNN model for gender/race identification: F1 score without repairing the data set 
% \end{enumerate}

\begin{table}[!tbp]
  \centering
  \caption{Demographic groups distribution in \feretdb}
  \label{tab:feretdb}
  \vspace{-3mm}
  \begin{tabular}{@{}lll|l@{}}
    \toprule
    & \textbf{Male} & \textbf{Female} & \textbf{Total} \\
    \midrule
    White & 331 & 229 & 560 \\
    Black & 21 & 19 & 40\\
    Asian & 80 & 47 & 127\\
    Hispanic & 11 & 8 & 19\\
    Middle Eastern & 9 & 1 & 10\\
    \midrule
    \textbf{Total} & 452 & 304 & 756\\
    \bottomrule
  \end{tabular}
\end{table}

\subsubsection{Performance metrics}
 In Task \ref{sec:exp:tasks:dataquality}, our performance metrics include the Quality Test Acceptance Rate and Data Distribution Test Acceptance Rate of the generated samples. For Tasks \ref{sec:exp:tasks:combselection}, our primary performance metric is the number of queries incurred by \fm, representing the cost of image generation. In Task \ref{sec:exp:tasks:humanevaluator}, our evaluation metrics are the Jaccard similarity of each algorithm's output with the ground truth. Finally, in our proof of concept, our metric focuses on the precision, recall and F1 score of the trained model on the test data set.

\subsection{Proof of Concept}\label{sec:exp:poc}

\begin{table*}[t]
    \centering
    \caption{Illustration of repairing lack of coverage and its effects on \texttt{\feretdb}}
    \label{tab:lackofcoverage}
    \vspace{-3mm}
    \begin{tabular}{lcccccccc}
        \toprule
        \multirow{2}{*}{\textbf{Ethnicity Groups}} & \multicolumn{4}{c}{\textbf{Classifier Performance on \texttt{\feretdb}}} & \multicolumn{4}{c}{\textbf{Classifier Performance on Repaired}} \\
        \cmidrule(lr){2-5} \cmidrule(lr){6-9}
        & \#Images & Precision & Recall & F1-Score & \#Images & Precision & Recall & F1-Score \\
        \midrule
        Overall          & 756 & 0.81 & 0.75 & 0.78 & 987 & 0.70 & 0.75 & 0.72 \\
        Black            & 40  & 0.19 & 0.22 & 0.16 & 100 & 0.48 & 0.56 & 0.52 \\
        Hispanic         & 19  & 0.50 & 0.17 & 0.25 & 100 & 0.62 & 0.36 & 0.45 \\
        Middle Eastern   & 10  & 0.00 & 0.00 & 0.00 & 100 & 0.20 & 0.41 & 0.27 \\
        \bottomrule
    \end{tabular}
\end{table*}

\begin{figure*}[!tb]
    \centering
    \begin{subfigure}{0.23\textwidth}
        \begin{tikzpicture}[scale=0.5]
          \begin{axis}[
            ybar,
            xlabel={Ethnicity Group},
            ylabel={F1-Disparity},
            legend pos=north west,
            bar width=15pt,
            symbolic x coords={Black,Hispanic,Middle Eastern},
            xtick=data,
            nodes near coords,
            nodes near coords align={vertical},
            enlarge x limits={0.2},
          ]
          \addplot[fill=americanrose, draw=red] table [x=group, y=disparity, col sep=comma] {f1-disparity-before.csv};
          \addlegendentry{Original \feretdb}
          \addplot[fill=airforceblue, draw=blue] table [x=group, y=disparity, col sep=comma] {f1-disparity-after.csv};
           \addlegendentry{Repaired \feretdb}
           \end{axis}
        \end{tikzpicture}
        \caption{F1-Disparity}
        \label{fig:f1-disparity}
    \end{subfigure}
    \hfill
    \begin{subfigure}{0.23\textwidth}
        \begin{tikzpicture}[scale=0.5]
          \begin{axis}[
            ybar,
            xlabel={Ethnicity Group},
            ylabel={Precision-Disparity},
            legend pos=north west,
            bar width=15pt,
            symbolic x coords={Black,Hispanic,Middle Eastern},
            xtick=data,
            nodes near coords,
            nodes near coords align={vertical},
            enlarge x limits={0.2},
          ]
          \addplot[fill=americanrose, draw=red] table [x=group, y=disparity, col sep=comma] {precision-disparity-before.csv};
          \addlegendentry{Original \feretdb}
          \addplot[fill=airforceblue, draw=blue]  table [x=group, y=disparity, col sep=comma] {precision-disparity-after.csv};
           \addlegendentry{Repaired \feretdb}
          \end{axis}
        \end{tikzpicture}
        \caption{Precision-Disparity}
        \label{fig:precision-disparity}
    \end{subfigure}
    \hfill
    \begin{subfigure}{0.23\textwidth}
        \begin{tikzpicture}[scale=0.5]
          \begin{axis}[
            ybar,
            xlabel={Ethnicity Group},
            ylabel={Recall-Disparity},
            legend pos=north west,
            bar width=15pt,
            symbolic x coords={Black,Hispanic,Middle Eastern},
            xtick=data,
            nodes near coords,
            nodes near coords align={vertical},
            enlarge x limits={0.2},
          ]
          \addplot[fill=americanrose, draw=red] table [x=group, y=disparity, col sep=comma] {recall-disparity-before.csv};
          \addlegendentry{Original \feretdb}
          \addplot[fill=airforceblue, draw=blue]  table [x=group, y=disparity, col sep=comma] {recall-disparity-after.csv};
           \addlegendentry{Repaired \feretdb}
          \end{axis}
        \end{tikzpicture}
        \caption{Recall-Disparity}
        \label{fig:recall-disparity}
    \end{subfigure}
    \hfill
    \begin{subfigure}{0.23\textwidth}
        \begin{tikzpicture}[scale=0.5]
          \begin{axis}[
            ybar,
            xlabel={Performance Metric},
            ylabel={Overall Performance Reduction},
            legend pos=north west,
            bar width=30pt,
            symbolic x coords={F1, Precision, Recall},
            xtick=data,
            nodes near coords,
            nodes near coords align={vertical},
            ymax=1,
            enlarge x limits={0.2},
          ]
        \addplot[fill=americanrose, draw=red] table [x=metric, y=cost, col sep=comma] {price-of-fairness.csv};
          \end{axis}
        \end{tikzpicture}
        \caption{Price of Fairness}
        \label{fig:l1-costs-plot}
    \end{subfigure}
    \caption{Unfairness (disparate performance) reduction for the uncovered groups in the \feretdb data set after data repair using \system, along with the price of fairness (overall performance reduction).}
    % \caption{Relationship between the Number of Repaired (a) L2 (b) L3 MUPs in the \utkface data set and the Total Cost Incurred by the \fm}
    \label{fig:disparities}
\end{figure*}
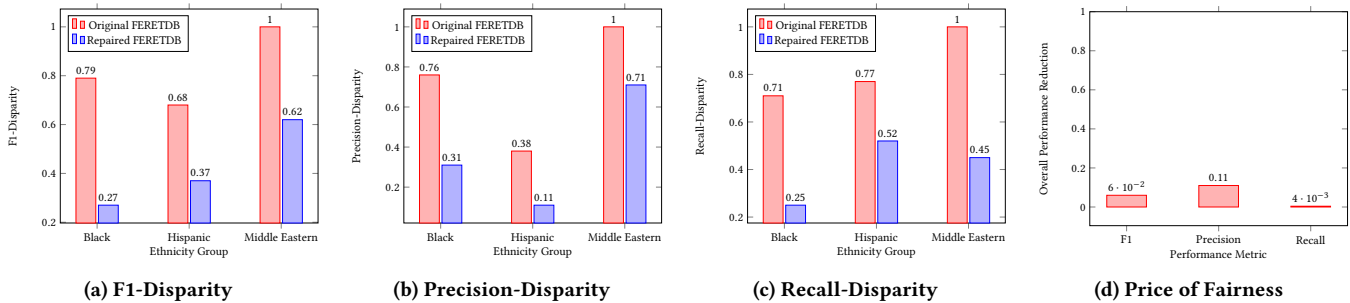

Our approach for resolving lack of coverage issues in a multi-modal data set is by augmenting it with {\em synthetically generated data}.
% The question we investigate in this section is {\em ``is this data-repair approach helpful?''}.
% Specifically, 
We start our experiments by investigating the \textit{feasibility}, \textit{effectiveness}, and \textit{efficiency} of this data-repair approach. 

To do so, we illustrate the impact of lack of coverage resolution using \system in a down-stream machine learning task.
We start by measuring the unfairness (in form of performance disparity) of a model trained on the original data set for under-represented groups.
% impacts of lack of coverage, in form of unfairness, on under-represented groups in an image data set.
Next, we repair the data set, using \system, and repeat the process to see if the unfairness issues reduce. 
% and then proceed to identify and resolve these issues using \system, examining the practical effectiveness of repairing the data set. 
Subsequently, we monitor the so-called ``price of fairness'', i.e., the  reduction in the overall performance as a result of unfairness reduction (data repair). %, exploring the costs associated with repairing the lack of coverage.

Our experiment employs the entire \feretdb data set as input, with detailed demographic group counts provided in Table~\ref{tab:feretdb}. 
While the data set has a reasonable coverage for both \at{male} and \at{female} genders, the racial groups \at{Black}, \at{Hispanic}, and \at{Middle eastern} are not covered (using the coverage threshold $\tau=100$).
We trained a race-predicting Convolutional Neural Network (CNN) model using this data set.
First, we observed an high overall performance of the model, with precision, recall, and F1-score being, 0.81, 0.75, and 0.78, respectively.
Moreover, the model shows similar performance on both (covered) genders. However, as reflected in Table~\ref{tab:lackofcoverage} (the ``classifier performance on \feretdb'' column), the model significantly under-performs for the uncovered groups. 
For example, while the overall F1-score is 78\%, it is as low as 16\%, 25\%, and 0\%, for the \at{Black}, \at{Hispanic}, and \at{Middle eastern} groups, respectively.
% While \feretdb demonstrates acceptable fairness rates for gender equality, significant disparities exist among ethnicity groups. We aim to showcase the real-world impact of this disparity. Following this, we train a Convolutional Neural Network (CNN) model to predict the race of each image in the data set, calculating precision, recall, and F1-score overall and for under-represented groups. 
% Table~\ref{tab:lackofcoverage} illustrates these performance metrics before any repair efforts. The model's performance for under-represented groups is notably poor.

To evaluate if resolving the lack of coverage for these groups using \system helps to reduce the gaps, we 
% To address these issues, we demonstrate that utilizing appropriate synthetic data can effectively resolve the lack of coverage. We 
employ it with the \greedy combination selection algorithm, Moderate mask delineation level, and the \linucb approach, to resolve the level-1 MUPs, i.e., the three uncovered racial groups. 
% with $\tau=100$ to rectify the lack of coverage. 
In total, \system issued {\em 307} queries to the foundation model,
% This results in the generation of a total of 307 images, 
of which {\em 231} pass both the quality and distribution tests. 
That is, {\bf 75\%} of the generated images passed the rejection sampling tests. We refer to the augmented data set as ``Repaired''.
Utilizing \dalle to generate 307 images incurred a total cost of \$4.91 (\$0.016 per image).
% The monetary cost of synthetic data generation presents a crucial aspect to consider. In this instance, utilizing \dalle to generate 307 images incurs a cost of \$4.91 (\$0.016 per image).

Next, we retrain the CNN using the Repaired data set.
% Subsequently, we repair the data set and retrain the CNN on the updated training data. 
Notably, the test data remains the same for both experiments and only contains real images. 
First, as expected, from Table~\ref{tab:lackofcoverage} (the ``classifier performance on Repaired'' column), one can notice a slight decrease on the overall performance of the model as a result of data augmentation.
On the other hand though, it is evident that {\em the performance of the model significantly increased for \underline{all} under-represented groups, across \underline{all} performance metrics.} For example, looking at the F1-scores, the performance improvement was more than 20\% in all cases.
% Table~\ref{tab:lackofcoverage} presents the improvements in precision, recall, and F1 score metrics for under-represented groups after repairing the data set, indicating a notable enhancement in performance metrics. 

Figures \ref{fig:f1-disparity}, \ref{fig:precision-disparity}, and \ref{fig:recall-disparity} show the model unfairness in form of Disparate Performance (F1-Disparity, Precision-Disparity, and Recall-Disparity) across the under-represented groups in the FERETDB data set, before and after the data repair.
The performance disparity for an under-represented group is computed as its performance ratio gap with the overall model performance. For example, if the overall performance of the model for a metric $p$ (e.g., F1-score) is $\rho_{all}$ and for a group $g$ is $\rho_g$, the unfairness is computed as
\[
p\text{-Disparity}(g) = \max\big(0, 1- \frac{\rho_g}{\rho_{all}}\big)
\]
% visually compares F1-Disparity, Precision-Disparity, and Recall-Disparity across various groups in the FERETDB data set before and after synthetic data augmentation. 
The figure demonstrates a clear reduction in disparities for all underrepresented groups after the repair process, which showcases the effectiveness of the data augmentation using \system.
For example, the F1-disparity for the \at{Black} group decreased from 79\% to 27\%.

\paragraph{Price of Fairness}
Due to the trade-offs between the model performance and fairness, improving fairness is usually associated with a reduction in the overall model performance, which is known as the {\em price of fairness}.
As we saw earlier, our data repair approach using \system could significantly reduce the model performance disparities for the under-represented groups.
This, however, comes at the cost of a slight model performance reduction. 
% The monetary cost of synthetic data generation presents a crucial aspect to consider. In this instance, utilizing \dalle to generate 307 images incurs a cost of \$4.91 (\$0.016 per image). However, evaluating the cost solely through direct generation expenses disregards potential downstream repercussions on model performance. 
Figure~\ref{fig:disparities} shows this cost, as the reduction in overall Precision, Recall, and F1-Score after data augmentation. The price of fairness, as reflected in various metrics, is modest compared to the substantial improvement in fairness achieved for under-represented groups.

\subsection{Performance Evaluation}

\subsubsection{Investigating the Influence of Mask Levels and Guide Image Selection on Quality Assessment \ref{sec:exp:tasks:dataquality}}

\begin{table*}[t]
    \centering
    \captionsetup{font=small} % Set smaller font size for the caption
    \caption{Performance of various Guide-selection algorithms}
    \label{tab:qtar}
    \resizebox{0.6\textwidth}{!}{%
    \begin{tabular}{p{2.5cm} p{2.5cm} cccc}
        \toprule
        \multirow{3}{*}{\textbf{\shortstack[l]{Guide\\Tuple\\Strategy}}} & \multirow{3}{*}{\textbf{\shortstack[c]{Mask\\Delineation\\Level}}} & 
        \multicolumn{2}{c}{\multirow{2}{*}{\textbf{\shortstack[c]{Quality Test\\ Acceptance Rate}}}} & 
        \multicolumn{2}{c}{\textbf{Data Distribution Test}} \\
        &  & & & \multicolumn{2}{c}{{\bf Acceptance Rate ($\boldsymbol{\nu = 0.3}$)}}\\
        \cmidrule(lr){3-4} \cmidrule(lr){5-6}
          &   & $\boldsymbol{\alpha = 0.1}$  & $\boldsymbol{\alpha = 0.4}$ &  \textbf{Linear} & \textbf{RBF}  \\
        \midrule
        No Guide 
        & -       & 0.90 & 0.81  & 0.53 & 0.44  \\
        \midrule
        \multirow{4}{*}{Random-Guide} 
        & Accurate & 0.69 & 0.51 & 0.70  & {\bf 0.74}  \\
        & Moderate & 0.85 & 0.70  & 0.70  & 0.70 \\
        & Imprecise & 0.90 & 0.70  & \textbf{0.73}  & 0.62 \\
        & \multicolumn{1}{r}{\textbf{Avg:}} & 0.81 & 0.64  & {0.71}  & {0.69} \\
        \midrule
        \multirow{4}{*}{Similar-Tuple} 
        & Accurate & 0.88 & 0.69  & 0.65  & 0.67  \\
        & Moderate & 0.90 & 0.75  & 0.64  & 0.58 \\
        & Imprecise & 0.85 & 0.68  & 0.65 & 0.53 \\
        & \multicolumn{1}{r}{\textbf{Avg:}} & 0.88 & 0.71 & 0.65 & 0.59  \\
        \midrule
        \multirow{4}{*}{LinUCB} 
        & Accurate & 0.90 &0.81  & 0.63 & 0.61 \\
        & Moderate & 0.91 &0.88  & 0.64 & 0.58  \\
        & Imprecise & \textbf{0.96} &\textbf{0.96}  & 0.63  & 0.54  \\
        & \multicolumn{1}{r}{\textbf{Avg:}} & \underline{0.92} &\underline{0.88} & 0.63  & 0.58 \\
        \bottomrule
    \end{tabular}%
    }
\end{table*}

\begin{figure}[!t]
\begin{tcolorbox}[
    colback=gray!10,
    colframe=gray!10,
    boxrule=0.5pt,
    rounded corners,
    boxsep=0pt,
    width=.48\textwidth
]
    \centering
    \includegraphics[width=.8\textwidth]{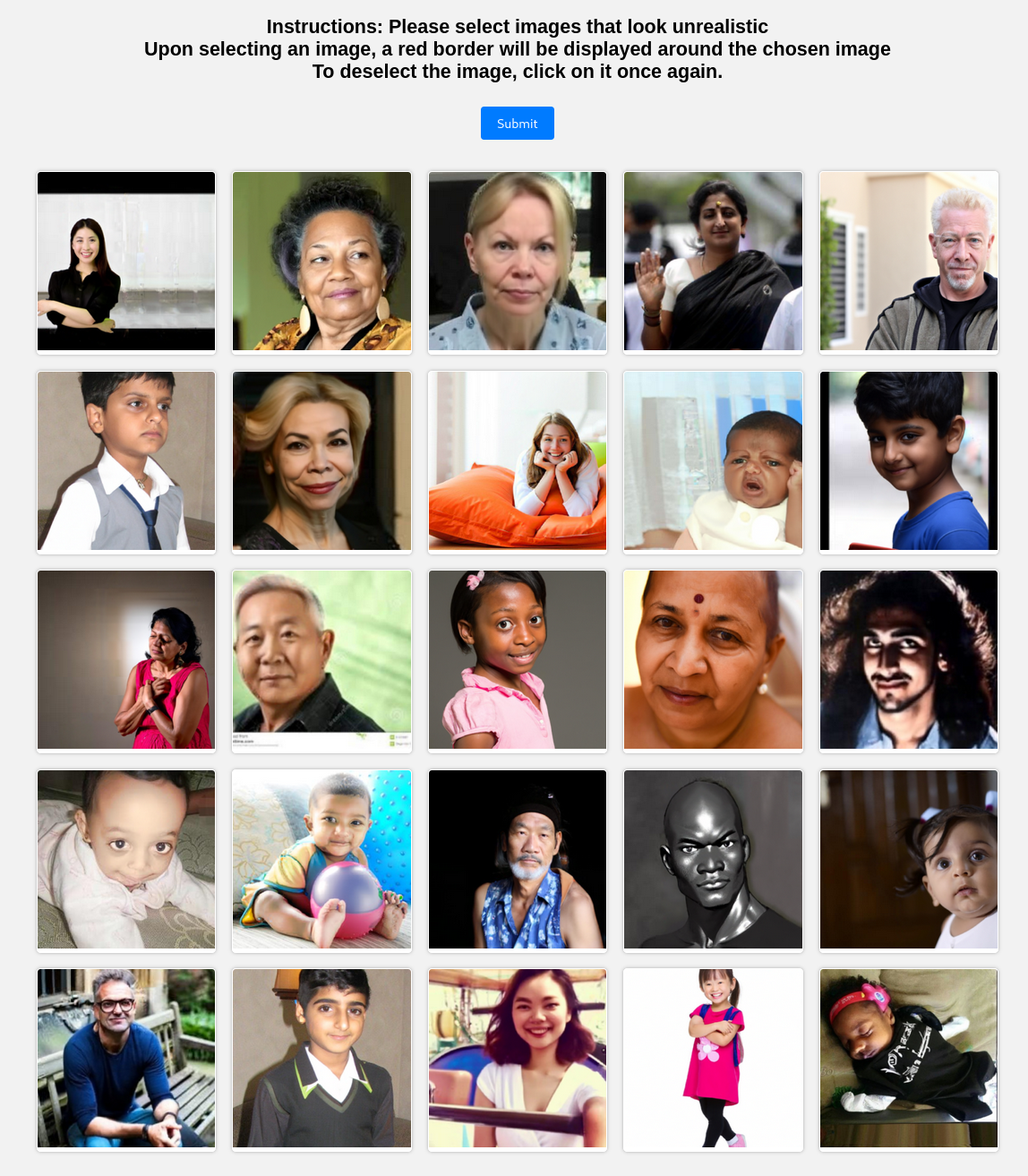}
    \vspace{-4mm}\caption{Sample Form Page Given to Participants}
    \label{fig:sample-reference}
\end{tcolorbox}
\vspace{-6mm}
\end{figure}

This study explores the impact of different mask delineation levels and guide tuple selection strategies on the performance of generated tuples in passing rejection sampling tests. A total of {\em 37 individuals} participated as the human evaluators for the quality evaluation test.
%We analyze their influence in various scenarios to understand how each combination affects the results.

To guarantee an inclusive evaluation, we intentionally constructed a challenging subset of the \utkface data set. This subset was designed to encompass Maximal Uncovered Patterns (MUPs) for all races and genders.  Within each age group, we introduced two distinct $\ell_3$ MUPs, each representing a different combination of gender and race (e.g., White male adult, Indian female adult). This approach ensures comprehensive coverage of various races, genders, and age groups during image generation. The exclusive use of $\ell_3$ MUPs guarantees that all experiments will generate identical combinations, effectively eliminating any potential randomness or variability in the results.

In total, we introduced 16 MUPs within the \utkface subset with $\tau=10$. Our objective was to resolve all MUPs in the subset for all possible combinations of guide tuple selection strategies and mask delineation levels. We then compared the Quality Test Acceptance Rate (QTAR) and Data Distribution Test Acceptance Rate (DDTAR) for each combination.

We generated a total of 831 images for these experiments and employed 27 human evaluators to assess their quality. Each evaluator received 200 images, presented in 8 pages of 25 images each. They were instructed to identify images that appear {\it unrealistic} to human beings. A sample of the evaluation form presented to participants is provided in Figure~\ref{fig:sample-reference}.

In a separate experiment, 
we employed 10 human evaluators for estimating the probability $p$ (Equation~\ref{eq:qualitytest}) that an evaluator labels a real image as realistic. 
To do so, we used the same forms as in Figure~\ref{fig:sample-reference}, except that this time all images were real, from the original data set. 
For \utkface, the probability $p$  was estimated as $0.86$.
% we repeat this process with 10 evaluators using real images from the tested data sets instead of synthetic data.\abol{what do you mean by tested data sets?} We calculate the probability of a real image being judged as realistic $p$ (\S~\ref{sec:test2}). 

For each setting, the Quality Test Acceptance Rate (QTAR) is defined as the number of images passing the quality test (\S~\ref{sec:test2}) divided by the total number of generated images. To explore the impact of evaluation stringency on QTAR, we calculated it for two significance levels, $\alpha=0.1$ and $\alpha=0.4$. A higher $\alpha$ value signifies a stricter acceptance policy, demanding a greater agreement among evaluators regarding image looking realistic. 

For $\alpha=0.4$, acceptance closely aligns with {\it unanimous agreement} among evaluators, whereas $\alpha=0.1$ approximates a {\it majority vote}, accepting images deemed realistic by over half of the evaluators. This distinction results in a trade-off between quality and quantity, with $\alpha=0.4$ yielding a smaller pool of images passing the test, with potentially higher overall quality. 
Table~\ref{tab:qtar} presents the calculated QTAR values for the designed \utkface subset under both significance levels.

The analysis of Quality Test Acceptance Rate (QTAR) reveals that the \linucb guide tuple selection strategy consistently outperforms No Guide, Similar-Tuple, and Random-Guide across both significance levels ($\alpha=0.1$ and $\alpha=0.4$). This performance gap widens further as the quality assessment becomes stricter (higher $\alpha$). Notably, images generated using \linucb exhibit demonstrably higher overall quality.

Further investigation into the interplay between guide-tuple selection strategies and mask delineation levels uncovers interesting trends. For both significance levels, Moderate and Imprecise mask delineation levels achieve superior QTAR compared to the Accurate level. This finding aligns with our initial expectations. Precisely cropping the foreground subject restricts the Foundation Model's creative freedom, potentially leading to unnatural entities generated to fill the cropped space. Conversely, Moderate and Imprecise levels provide greater flexibility, allowing the Foundation Model to generate new objects more naturally and potentially contributing to improved image quality.

Next, we move to the data distribution test (\S~\ref{sec:test1}). DDTAR, defined as the proportion of images passing the test, assesses how well the generated images adhere to the underlying data distribution. We use {\sc MobileNetV3}\cite{howard2019searching} as the embedder and OCSVM ($\nu=0.3$) with two kernels (RBF and Linear) for training. The choice of the kernel impacts the performance of the OCSVM, with the RBF kernel often capturing complex relationships in the data, while the Linear kernel assumes linearity. 
% Additionally, we perform each experiment with two distinct values for $\nu$. A higher $\nu$ leads to a smaller and tighter boundary, potentially excluding more normal data points as outliers, while a lower $\nu$ results in a larger and looser boundary, tolerating more outliers but risking misclassifying normal data.

% As anticipated, the No-Guide strategy suffers significant performance decline under tight boundaries ($\nu=0.3$). This drop in Data Distribution Test Acceptance Rate (DDTAR) – 37.5\% on average, compared to 26\% for Random-Guide, 27\% for Similar-Tuple, and 25.5\% for LinUCB – suggests an actual {\em distribution drift} in images generated by No-Guide. This highlights the importance of guiding the generation process toward the target distribution using Guide Tuples.

% strategy suffers significant performance decline under tight boundaries ($\nu=0.3$). This drop in Data Distribution Test Acceptance Rate (DDTAR) – 37.5\% on average, compared to 26\% for Random-Guide, 27\% for Similar-Tuple, and 25.5\% for LinUCB – suggests an actual {\em distribution drift} in images generated by No-Guide. This highlights the importance of guiding the generation process toward the target distribution using Guide Tuples.

The No-Guide strategy does not provide a guide for the image generation process, leaving the details to the imagination of the foundation model. As a result, it is anticipated that a larger portion of the images generated with this strategy should fail the data distribution test. This is confirmed in Table~\ref{tab:qtar}, where around half of the images generated with this strategy (using either of the two kernels) could not pass the test.
On the other hand, the Random-Guide strategy is focused on following the data distribution. Therefore, viewing the images in the data set as the random iid samples from its underlying distribution, it draws a random image from the data set and uses it as the guide, irrespective of the description of the image to be generated. This approach, while has a smaller chance of passing the quality test, is expected to have the highest chance of passing the data distribution test.
Our findings in Table~\ref{tab:qtar} are consistent with this expectation. 
% Another important observation is that the Random-Guide strategy outperforms Similar-Tuple and \linucb on average across all boundary settings. This can be attributed to its focus on preserving the underlying distribution, unlike the other strategies, Random-Guide draws from the largest pool of potential guide tuples, treating each one equally. While this approach lacks the targeted selection of Similar-Tuple and \linucb, it effectively maintains the desired data distribution, even with tight boundaries. 
While the Random-Guide strategy outperforms \linucb and the Similar-Tuple strategy on the data distribution test, both of these strategies still demonstrate acceptable performance. 
Regarding mask delineation levels, Accurate delineation exhibits marginally higher DDTAR compared to Moderate and Imprecise levels. This aligns with our expectations, as stricter cropping likely helps constrain the generated images to closer proximity to the original data distribution. However, the performance difference is relatively small, suggesting that the advantages of Moderate and Imprecise delineation in terms of naturalness and flexibility outweighs the slight decrease in distribution adherence for tight boundaries.

Overall, since {\em \linucb outperforms the other approaches on the quality evaluation test (which involves human evaluators) and shows an acceptable performance on the data distribution test}, it is the preferred approach for guide-selection.

\subsubsection{\greedy Combination Selection Algorithm \ref{sec:exp:tasks:combselection}}
\begin{figure*}[!tb]
    \centering
    \begin{subfigure}{0.23\textwidth}
        \begin{tikzpicture}[scale=0.5]
          \begin{axis}[
            scatter,
            xlabel={\#$\ell_2$ MUPS Repaired},
            ylabel={\#Images need to be generated},
            legend pos=north west,
            bar width=2pt,
          ]
          \addplot[mark = o, mark size=4pt] table [x=l2r, y=total, col sep=comma] {greedy-l2-200.csv};
          \addlegendentry{\greedy}

          \addplot[mark = x, mark size=5pt] table [x=l2r, y=total, col sep=comma] {random-l2-200.csv};
           \addlegendentry{Random}

           \addplot[mark = +, mark size=5pt] table [x=l2r, y=total, col sep=comma] {bestcomb-l2-200.csv};
           \addlegendentry{Min-Gap}
          \end{axis}
        \end{tikzpicture}
        \caption{$\tau=200$}
    \end{subfigure}
    \hfill
    \begin{subfigure}{0.23\textwidth}
        \begin{tikzpicture}[scale=0.5]
          \begin{axis}[
            scatter,
            xlabel={\#$\ell_2$ MUPS Repaired},
            ylabel={\#Images need to be generated},
            legend pos=north west,
            bar width=2pt,
          ]
          \addplot[mark = o, mark size=4pt] table [x=l2r, y=total, col sep=comma] {greedy-l2-350.csv};
          \addlegendentry{\greedy }

          \addplot[mark = x, mark size=5pt] table [x=l2r, y=total, col sep=comma] {random-l2-350.csv};
           \addlegendentry{Random}

           \addplot[mark = +, mark size=5pt] table [x=l2r, y=total, col sep=comma] {bestcomb-l2-350.csv};
           \addlegendentry{Min-Gap}

          \end{axis}
        \end{tikzpicture}
        \caption{$\tau=350$}
    \end{subfigure}
    \hfill
    \begin{subfigure}{0.23\textwidth}
        \begin{tikzpicture}[scale=0.5]
          \begin{axis}[
            scatter,
            xlabel={\#$\ell_1$ MUPS Repaired},
            ylabel={\#Images need to be generated},
            legend pos=north west,
            bar width=8pt,
          ]
          \addplot[mark = o, mark size=4pt] table [x=l1r, y=total, col sep=comma] {greedy-l1-1000.csv};
          \addlegendentry{\greedy}

          \addplot[mark = x, mark size=5pt] table [x=l1r, y=total, col sep=comma] {random-l1-1000.csv};
           \addlegendentry{Random}

           \addplot[mark = +, mark size=5pt] table [x=l1r, y=total, col sep=comma] {bestcomb-l1-1000.csv};
           \addlegendentry{Min-Gap}

          \end{axis}
        \end{tikzpicture}
        \caption{$\tau=1000$}
    \end{subfigure}
    \hfill
    \begin{subfigure}{0.23\textwidth}
        \begin{tikzpicture}[scale=0.5]
          \begin{axis}[
            scatter,
            xlabel={\#$\ell_1$ MUPS Repaired},
            ylabel={\#Images need to be generated},
            legend pos=north west,
            bar width=6pt,
            xtick=data
          ]
          \addplot[mark = o, mark size=4pt] table [x=l1r, y=total, col sep=comma] {greedy-l1-2000.csv};
          \addlegendentry{\greedy }

          \addplot[mark = x, mark size=5pt] table [x=l1r, y=total, col sep=comma] {random-l1-2000.csv};
           \addlegendentry{Random}

           \addplot[mark = +, mark size=5pt] table [x=l1r, y=total, col sep=comma] {bestcomb-l1-2000.csv};
           \addlegendentry{Min-Gap}

          \end{axis}
        \end{tikzpicture}
        \caption{$\tau=2000$}
    \end{subfigure}

    \caption{Comparison of cost across various thresholds for \greedy, Random, and Min-Gap combination selection algorithms}
    \label{fig:combined-costs-plots}
\end{figure*}
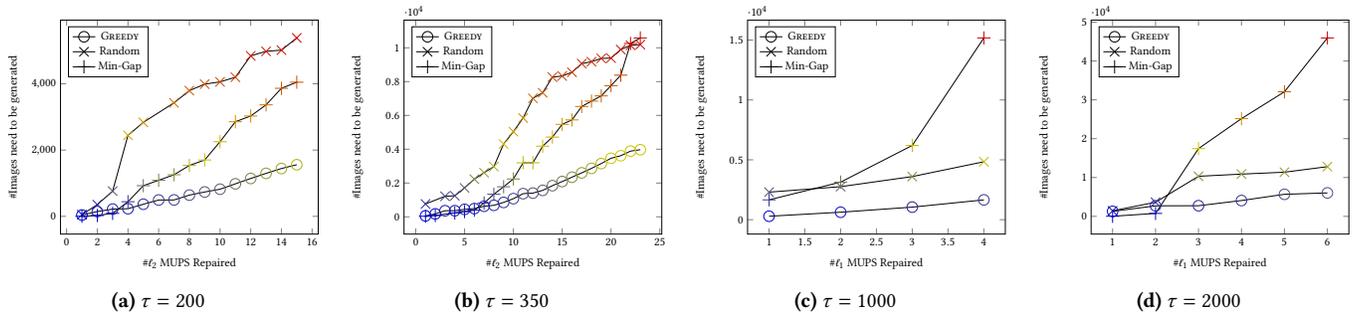

In this experiment, we study the impact of employing the \greedy algorithm for selecting the next combination to generate tuples. The entire \utkface data set serves as input for our investigation, where we analyze the cost of repairing all $\ell_1$ or $\ell_2$ MUPs for different thresholds using various combination selection algorithms.  As a baseline, we employ the Random selection algorithm, which randomly chooses a combination in each iteration without considering the MUPs' status. 
We also introduce the {\sc Min-Gap} algorithm, which given list of MUPs, first identifies the MUP which has smallest gap $\delta$ from treshold $\tau$, then chooses a combination that satisfies this MUP and generates $\delta$ tuples to satisfy that MUP. Unlike the \greedy algorithm, the {\sc Min-Gap} Algorithm only focuses on the distance from the threshold, disregarding the number of MUPs hit and MUPs level. We conduct experiments with four distinct values for $\tau$ and monitor the total number of images each algorithm requires to add to the data set for resolving the MUPs.
% number of queries to the Foundation Model \fm. The simulated \fm serves as a {\it perfect} Image Generator, ensuring that its generated results consistently pass quality and distribution tests. 
% The associated cost with each strategy is indicator of number of images that should be generated and pass the rejection sampling tests.

% In each experiment, We will focus on resolving all lowest level MUPs. 
For $\tau=200$ and $\tau=350$, all MUPs are at levels $\ell_2$ and $\ell_3$. 
In these experiments we the goal is to resolve $\ell_2$ MUPs.
% The experiment continues until all $\ell_2$ MUPs are resolved. 
For $\tau=1000$ and $\tau=2000$, where $\ell_1, \ell_2$, and $\ell_3$ MUPs are present, our focus is on resolving all $\ell_1$ MUPs.

Figure~\ref{fig:combined-costs-plots} shows the associated cost (total number of images to be generated) for each strategy in different scenarios.
We can observe that in all cases \greedy algorithm significantly outperforms both Random and {\sc Min-Gap} baselines. The Gap even becomes more noticeable when trying to resolve lower level MUPs. In addition, the {\sc Min-Gap} strategy performs better than Random for $\ell_2$ MUPs  but significantly worse when attempting to satisfy the $\ell_1$ MUPs. This is due to the larger pool of MUPs in higher thresholds, as the Min-Gap algorithm may choose numerous irrelevant MUPs to satisfy, leaving $\ell_1$ MUPs unsatisfied for subsequent iterations.

\subsubsection{Quality Assessment Tools Comparison \ref{sec:exp:tasks:humanevaluator}}

% \begin{table}[!bt]
%   \centering
%   \caption{Jaccard similarity of the output of each image quality assessment algorithm with the ground truth set}
%   \label{tab:humanevaluator}
%   \vspace{-3mm}
%   \begin{tabular}{@{}lll@{}}
%     \toprule
%     & \textbf{Metric Mode} & \textbf{$J_s$} \\
%     \midrule
%     NIQE & NR & 0.127 \\
%     BRISQUE & NR & 0.068 \\
%     NIMA & NR & 0.068 \\
%     % VIF & FR & \\
%     \bottomrule
%   \end{tabular}
% \end{table}
\begin{table}[!bt]
  \centering
  \caption{Jaccard similarity of the output of each image quality assessment algorithm with the ground truth set}
  \label{tab:humanevaluator}
  \vspace{-3mm}
  \begin{tabular}{@{}ll@{}}
    \toprule
    \textbf{Quality Assessment Algorithm} & \textbf{$J_s$} \\
    \midrule
    NIQE &  0.127 \\
    BRISQUE & 0.068 \\
    NIMA & 0.068 \\
    % VIF & FR & \\
    \bottomrule
  \end{tabular}
\end{table}
In this experiment we investigate the performance of multiple image quality assessment algorithms by comparing their outputs with human evaluator ground truth. The objective is to explore the feasibility of transitioning from human evaluators to automated tools. Before continuing, we would like to underscore that the purpose of this step is not similar to ``fake image'' detectors, such as in \cite{kiruthika2023image,liu2023detection}, to distinguish between synthesized and authentic images. Note that, the evaluated images in this step are indeed machine-generated. %Instead, our focus is solely on discerning the natural quality of machine-generated images.

Image quality evaluation can be accomplished using established techniques such as Blind/Referenceless Image Spatial Quality Evaluator ({\sc BRISQUE}) ~\cite{mittal2012no},
Neural Image Assessment ({\sc NIMA}) \cite{talebi2018nima}, or Natural Image Quality Evaluator ({\sc NIQE})~\cite{mittal2012making}. 
For instance, given an image to evaluate, NIQE first generates a masked version of the image using the weighted local mean and contrast of the image in the vicinity of each pixel. It then creates a feature matrix for the image using a Multivariate Gaussian (MVG) density model. The mean and covariance of the feature matrix (compared with those of a training set) are then used to assign a quality measure and classifying the image as ``good quality'' or not. Each algorithm returns a score, and an acceptance threshold $\tau_c$ is defined for each tool. We fine-tune $\tau_c$ based on the actual number of rejected images.

In this experiment, a dataset of 271 synthetic images, generated using guide images from \utkface and varying mask delineation levels is subjected to 7 human evaluators. Each evaluator is asked to identify images that ``do not look realistic.'' with an average exposure of each image to more than five evaluators. We use labeling procedure outlined in \S~\ref{sec:test2} with significance level $\alpha = 0.1$ to label images. Out of the total, 27 images are marked as "{\it unacceptable}" while the remaining images are labeled as acceptable. 

We subjected the ground truth dataset to BRISQUE ~\cite{mittal2012no}, NIMA ~\cite{talebi2018nima} and NIQE ~\cite{mittal2012making}.
For each algorithm, we calibrated acceptance thresholds to reject precisely 27 images, aligning with the number identified as unacceptable by human evaluators. We compared the sets of rejected images between each algorithm and the ground truth using Jaccard Similarity. This metric quantifies the degree of overlap between the sets, with values closer to 1 indicating stronger agreement. Table~\ref{tab:humanevaluator} presents the Jaccard Similarity scores for each algorithm. Results reveal that none of the assessed algorithms achieved satisfactory performance in reliably isolating unrealistic images.

% Thus, a human-in-the-loop approach is adopted, where a grid of produced images is presented to human evaluators, who identify low-quality images.

% \abol{add necessary missing details about this step as needed}

% \abol{don't forget mention the tradeoff of parallelization of multiple image production with UCB at the end of section 4}

%% file: related.tex
\section{Related Work}\label{sec:related}
%  Resources that I've found useful, particularly, their related work section is very similar to what we need 

% https://dl.acm.org/doi/pdf/10.1145/3543507.3583341
% http://vldb.org/pvldb/vol14/p2519-nargesian.pdf
% https://dl.acm.org/doi/pdf/10.1145/3514221.3517841

% Topics 
% Responsible Data Science
% Bias and Representativeness in Data (Data Coverage)
% State of Foundation Model (LLM) in Data management community  

% \paragraph{Responsible Data Science}
% paraphrase from data engineering bulletin 
While data bias has been a long-standing concern in the statistical community~\cite{neyman1936contributions}, social data presents unique challenges due to its inherent complexity and sensitivity~\cite{olteanu2019social,fairmlbook,barocas2016big,jk2019bias,drosou2017diversity}. Issues of diversity and representativeness have been studied across various disciplines, including social science~\cite{berrey2015enigma, dobbin2016diversity,simpson1949measurement}, political science~\cite{surowiecki2005wisdom}, and information retrieval~\cite{agrawal2009diversifying}. Efforts to trace machine bias back to its sources involve identifying different types~\cite{mehrabi2021survey, olteanu2019social,friedman1996bias} and sources~\cite{torralba2011unbiased,crawford2013hidden,diakopoulos2015algorithmic} of biases in data. Existing work to meet {\it responsible data} requirements~\cite{nargesian2022responsible} extend throughout various stages of the data analysis pipeline, including data annotation~\cite{li2020towards,lazier2023fairness}, data cleaning and repair~\cite{SalimiRHS19,tae2019data,salimi2020database}, data imputation~\cite{martinez2019fairness}, entity resolution~\cite{shahbazi2023through,fanourakis2023fairer}, and data integration~\cite{nargesian2022responsible,nargesian2021tailoring}.

\paragraph{Data Coverage} 
% paraphrase from data engineering bulletin 
The notion of data coverage has been proposed for detecting under-representation issues in a data set~\cite{shahbazi2023representation}.
Related work in this area can be divided into (a) lack of coverage detection and (b) lack of coverage resolution.
Coverage detection in tabular data has been studied for both discrete~\cite{asudeh2019assessing} and continuous~\cite{asudeh2021coverage} attributes, whether in single or multiple relations~\cite{lin2020identifying}, and recently for image data set~\cite{mousavi2024data}.
Efforts for addressing lack of coverage without additional data collection include query rewriting~\cite{accinelli2020coverage, accinelli2021impact,shetiya2022fairness} and generating lack of representation warning~\cite{shahbazi2022data}. 
On the other hand, data tailoring \cite{nargesian2021tailoring} has been proposed to integrate additional data from a data lake to resolve under-representation issues. Differently, for tabular data, \cite{sharma2020data,iosifidis2018dealing,celis2020data} partially alter duplicates of existing tuples, or generate synthetic entries from existing data to resolve lack of coverage.
% include rewriting queries to resolve representation bias~\cite{accinelli2020coverage, accinelli2021impact,shetiya2022fairness}, minimizing additional data collection costs~\cite{asudeh2019assessing,tae2021slice}, and employing data augmentation methods by introducing partially altered duplicates of existing tuples or generating synthetic entries from existing data~\cite{sharma2020data,iosifidis2018dealing,celis2020data}. Another strategy involves using data integration techniques to consolidate diverse sources into a single dataset~\cite{nargesian2021tailoring}. As an alternative approach to measure representation bias, the concept of representation rate~\cite{celis2020data} (also known as equal base rate~\cite{kleinberg2016inherent}) is introduced. In contrast to coverage, representation rate is more restrictive as it requires nearly equal ratios from different groups. 
For a comprehensive survey on representation bias in data, refer to \cite{shahbazi2023representation}.

\paragraph{Foundation Models for Data Management}
% Main resource: https://github.com/uncbiag/Awesome-Foundation-Models
With recent advancements in large language models~\cite{devlin2018bert, brown2020language, touvron2023llama} and foundation models
, those have been widely used in various research communities for tasks such as Code Generation \cite{chen2021evaluating}, Synthetic Image Generation \cite{ramesh2022hierarchical}, and Video Generation \cite{bar2024lumiere}.
The current state of utilizing these models in the data management community reflects a growing recognition of their potential and challenges. 
Some of the recent work within for utilizing generative AI for data management problems are as following.
LLMs have shown an extraordinary performance for answering natural language queries~\cite{trummer2023demonstrating,chang2023prompt,ye2023large}. 
Particularly, {\sc ThalamusDB} enables answering complex natural language queries on multi-modal data~\cite{jo2023demonstration}.
LLMs and founcation models have also been utilized for challenging tasks such as
data set search~\cite{trummer2023demonstrating},
predicting data correlations~\cite{trummer2023can},
data-lake profiling~\cite{arora2023language},
and anomaly detection in time-series~\cite{chen2023imdiffusion}, to name a few.
Still, to the best of our knowledge, none of the existing work has utilized foundation models for fairness-aware multi-modal data augmentation.

% Regarding the Fairness in Foundation models, there have been much less research done,  \cite{hua2023up5} proposes a method to reduce unfair recommendation results in LLMs. 

%% file: proofs.tex
\section{Proofs}\label{sec:proofs}

\noindent{\sc Lemma}~\ref{lem:nphardness}.
{\em
\cs is \np-hard.
}

\begin{proof}
We prove the \np-hardness of \cs using the reduction from {\em Vertex-cover} (\vc).
In the proof, we assume the combinations $C$ to choose from are specified as part of the input to \cs.

The input to \vc is $G(V,E)$, a graph with the set of vertices $V=\{v_1,\cdots,v_{n'}\}$ and the set of edges $E=\{e_1,\cdots,e_{m'}\}$.
Then, given a value $k$, the decision version of \vc determines if there exists a subset $V'\subseteq V$ of size (at most) $k$, such that all edges are covered, i.e., $\forall e_i=(u,v)\in E$, $\{u,v\}\cap V' \neq \emptyset$.

Given a target value $k$, the decision version of \cs problem determines if there exists an assignment to $\sigma$-values such that $\sum_{c_i\in C} \sigma_i\leq k$, while all MUPs $M\in\mathcal{M}^*$ are resolved.
Given an instance of \vc, we reduce it to \cs in $P$-time as following:
\begin{itemize}[leftmargin=*]
\item For each edge $e_i$, add a binary attribute $x_i$ to the attributes of interest. That is $\mathbf{x}=\{e_1,\cdots,e_{n'}\}$.
\item Consider the binary matrix $C$, with $m'$ rows and the columns $e_1$ to $e_{n'}$. 
Add $m'$ rows $v_1$ to $v_{m'}$, where the $i$-th row corresponds with the vector $v_i$.
For each edge $e_i=(v_j,v_k)$, set the cells $C_{i,j}$ and $C_{i,k}$ as 1 and all other cells in column $e_i$ as 0.
Therefore, in sum of the values on each column is exactly 2.
The rows of matrix $C$ represent the combinations to choose from. In other words, for each vector $v_i\in V$, a binary value combination $C_i$ is added, where (only) the cells corresponding to its edges are one.
\item For each edge $e_i$ add the level-1 MUP $M_i = X\cdots X1X\cdots$, where the $i$-th element is 1 and all others are unspecified. Set $\delta(M_i)=1$.
\item Set the target value in the decision version of \cs as $k$. 
% For every element $u_i\in \cU$, we add two elements $u_i^0$ and $u_i^1$ to $\cU'$, where $\cC(u_i^j)=j$. Note that $|\cU'| = 2n$.
% \item For every set $\cS_i$ we add the set $\cS'_i$. For each element $u_i\in\cS_i$, we add $u_i^0$ and $u_i^1$ to $\cS'_i$.
\end{itemize}
Now, the output to the \vc problem is yes, i.e., a vertex cover of size at most $k$ exists, if and only if the output to the \cs problem is yes.
As a result, since \vc is \np-complete, \cs is \np-hard, unless {\sc P}=\np.
\end{proof}

\vspace{3mm}
\noindent{\sc Theorem}~\ref{th:greedy}.
{\em
    The approximation ratio of the \greedy approach is $\log(\eta)$, where $\eta=\sum_{M\in\mathcal{M}^*}\delta(M)$.
}

\begin{proof}
    At every iteration, the \greedy algorithm selects a combination that contributes the most in reducing the coverage gap for a subset of remaining MUPs in $\mathcal{M}^*$.
    In order to prove the approximation ratio, we assign a price of 1 for each time a combination $c$ is selected, i.e., for each increase of 1 in the count of $\sigma(c)$.
    This price is uniformly distributed across the remaining MUPs that $c$ match to.
    Now, let us define the variable $e[i,j]$ that shows the price paid by the algorithm to reduce the gap of $M_i$ for the $j$-th time (each time the gap is reduced by one).
    For example, if during the current iteration, the selected combination $c$ matches the remaining MUPs $\{M_2,M_5,M_7\}$, while this is the first match for $M_2$, the 6-th match for $M_5$, and the second match for $M_7$, we set $e[2,1]=1/3$, $e[5,6]=1/3$, and $e[7,3]=1/3$.
    
    Let us consider the ordering of $e[i,j]$ values from the smallest to largest and let $\eta=\sum_{M\in\mathcal{M}^*}\delta(M)$.
    Let the ordered list be indexed as $\mathbf{e}=\langle e_1, e_2,\cdots,e_\eta\rangle$.

    Now, let \opt be the total sum of the optimal combination-counts (i.e., $\min \sum \sigma_i$) for resolving the MUPs $\mathcal{M}^*$. In other words, \opt is the total price that the optimal algorithm pays to fills the gaps for all MUPs in $\mathcal{M}^*$.
    For each $e_i=e[k,j]$ in the ordered list, let $e_i^o$ be the price assigned to match $M_k$ for the $j$-th time.
    Uniformly distributing \opt across $\langle e_i, e_{i+1},\cdots,e_\eta\rangle$, we know that
    \[
    e_i^o\leq \frac{\opt}{\eta-i+1}
    \]
    Also, since at each iteration, greedy selects the combination that matches the maximum number of remaining MUPs
    \[
    e_i\leq e_i^o
    \]
    Let {\sc A} be the total sum of the combination-counts by the \greedy algorithm. Since {\sc A} is distributed between the elements of $\mathbf{e}$,

    \begin{align*}
        &A = \sum_{i=1}^\eta e_i \leq \sum_{i=1}^\eta e_i^o\leq \sum_{i=1}^\eta \frac{\opt}{\eta-i+1} = \opt \sum_{i=1}^\eta \frac{1}{\eta-i+1} \\
        &\sum_{i=1}^\eta \frac{1}{\eta-i+1} =\sum_{i=1}^\eta \frac{1}{i} = H_\eta\\\vspace{2mm}
        \Rightarrow \quad& A \leq \opt ~H_\eta = \opt ~\Theta\big(\log (\eta)\big) \\\vspace{2mm}
        \Rightarrow \quad& \frac{A}{\opt} \leq \Theta\big(\log (\eta)\big)
    \end{align*}
\end{proof}

%% file: conclusion.tex
\section{Conclusion}\label{sec:conclusion}
In this paper, we introduced \system for fairness-aware data augmentation to reduce the under-representation of minority groups.
Motivated by the recent advancements in the foundation models, our system efficiently utilizes them for data-repair with minimum addition of synthetically generated data, while ensuring the augmented data is of high quality and follows the underlying data distribution. Our experiment results demonstrated the effectiveness of our data-repair approach in reducing the unfairness of a down-stream task.
This motivates the future work to extend the scope of fairness-aware data augmentation to other settings such as natural language data and graph data, using large language models. 